%% file: main_en.tex
\PassOptionsToPackage{numbers,sort&compress}{natbib}
\PassOptionsToPackage{table,dvipsnames}{xcolor}
\documentclass[]{style/company_light}

\usepackage{hyperref}
\usepackage{url}
\usepackage{booktabs}
\usepackage{amsfonts}
\usepackage{amsmath}
\usepackage{nicefrac}
\usepackage{microtype}
\usepackage{graphicx}
\usepackage{subcaption}

\usepackage{multirow}
\usepackage{enumerate}
\usepackage{enumitem}
\usepackage[most]{tcolorbox}
\usepackage{adjustbox}
\usepackage{array}
\usepackage{amssymb}

\newcommand{\yes}{\textcolor{ForestGreen}{$\checkmark$}}
\newcommand{\no}{\textcolor{red}{$\times$}}
\newcommand{\partialyes}{\textcolor{red}{$\checkmark$}}

\definecolor{todored}{RGB}{204,0,0}

\newcommand{\bench}{SWE~Refactor~Bench}
\newcommand{\stA}{State~A}
\newcommand{\stB}{State~B}

\newcommand{\keyclaim}[1]{{\bfseries #1}}

\newcommand{\sI}{Migration Audit}
\newcommand{\sII}{Behavioural Tests}
\newcommand{\sIII}{Agentic Verification}

\title{\fontsize{16.5pt}{21pt}\selectfont\baselineskip=21pt\relax
SWE Refactor Bench: Can Coding Agents Complete a\\
Long-Horizon, Whole-Repository Stack Migration?\par}
\author{
    \footnotesize Deyao Hong$^{*\dagger}$, Yizhe Chi$^{*}$, Wenyi Li$^{*}$, Xiaoqiu Wang, Mingju Gao,
    Kaisen Yang, Bingxiang He, Youjie Zheng, Calvin Xiao, Qinhuai Na$^{\ddagger}$
}
\renewcommand\affiliationformat[2][]{\makebox[\linewidth][c]{\small\bfseries #2}}
\affiliation{Navers Lab, Einsia.AI\quad Tsinghua University}
\renewcommand\contributionformat[2][]{%
  \vskip 0.15cm
  \makebox[\linewidth][c]{\footnotesize\color{gray}#2}%
}
\contribution{$^{*}$Equal Contribution\quad$^{\dagger}$Project Lead\quad$^{\ddagger}$Corresponding Author}

\input{Sections_en/0_abstract}

\metadata[Date]{\today}
\metadata[Homepage]{\url{https://lab.einsia.ai/swe-refactor-bench}}

\begin{document}

\maketitle

\input{Sections_en/1_introduction}
\input{Sections_en/2_benchmark}
\input{Sections_en/3_experiments}
\input{Sections_en/4_relatedwork}

\input{Sections_en/5_conclusion}

% ── References ───────────────────────────────────────────────────────────────

\clearpage
\newpage
\bibliographystyle{assets/plainnat}
\bibliography{ref}

% ── Appendices ───────────────────────────────────────────────────────────────

\clearpage
\newpage
\appendix

% Three parts: how a task is built (with the lang01 walk-through), the per-task
% breakdown of the results, and the task catalogue.
\input{Sections_en/A_construction}
\input{Sections_en/B_per_task}
\input{Sections_en/C_catalog}

\end{document}

%% file: Sections_en/0_abstract.tex
\abstract{
Modern software systems accumulate technical debt over decades of development, which makes migration expensive and largely manual.
As coding agents become increasingly capable at bug fixing, can they autonomously perform such migrations? 
Existing benchmarks cannot answer this question because they evaluate only behavioural correctness, not whether the migration actually occurred. 
This leads an easy hack: agents copy the original implementation to make tests pass. We call this \emph{Blindness}.
To address this problem, we introduce \textbf{\bench{}}, a benchmark comprising 20 whole-repository migrations, covering 4 kinds of technical debt. 
A three-stage evaluation protocol measures both migration completeness and behavioural correctness.
(1) \sI{} verifies that the migration occurred. 
(2) \sII{} measure correctness with a fixed test suite. 
(3) \sIII{} uses 6 independent coding agents to generate targeted tests for hidden behavioural differences.
Across 520 runs from 8 frontier models and 26 model-effort configurations,
only 28 of 520 runs ($5.4\%$) pass all three stages, 13 of the 20 tasks receive no accepted solution, and the best model (claude-opus-5)  scores $47.0/100$.
Migration completeness and behavioural correctness are distinct abilities: a few runs preserve behaviour by skipping the migration and are stopped at \sI{}; most attempt it and break behaviour, and are stopped at \sII{}.
Agents cannot deliver a perfect migration: among the 340 runs that pass \sI{}, $58\%$ reach $99\%$ of the fixed checks, yet only $26\%$ reach $100\%$.
Agent capability differs across migration categories: agents score $31.4$ on build toolchain rewrites but only $5.6$ on language rewrites.
Together, these findings position \bench{} as a rigorous testbed for developing coding agents for reliable whole-repository migrations.
}

%% file: Sections_en/1_introduction.tex
\section{Introduction}
\label{sec:introduction}

\begin{figure}[t]
  \centering
  \includegraphics[width=0.98\textwidth]{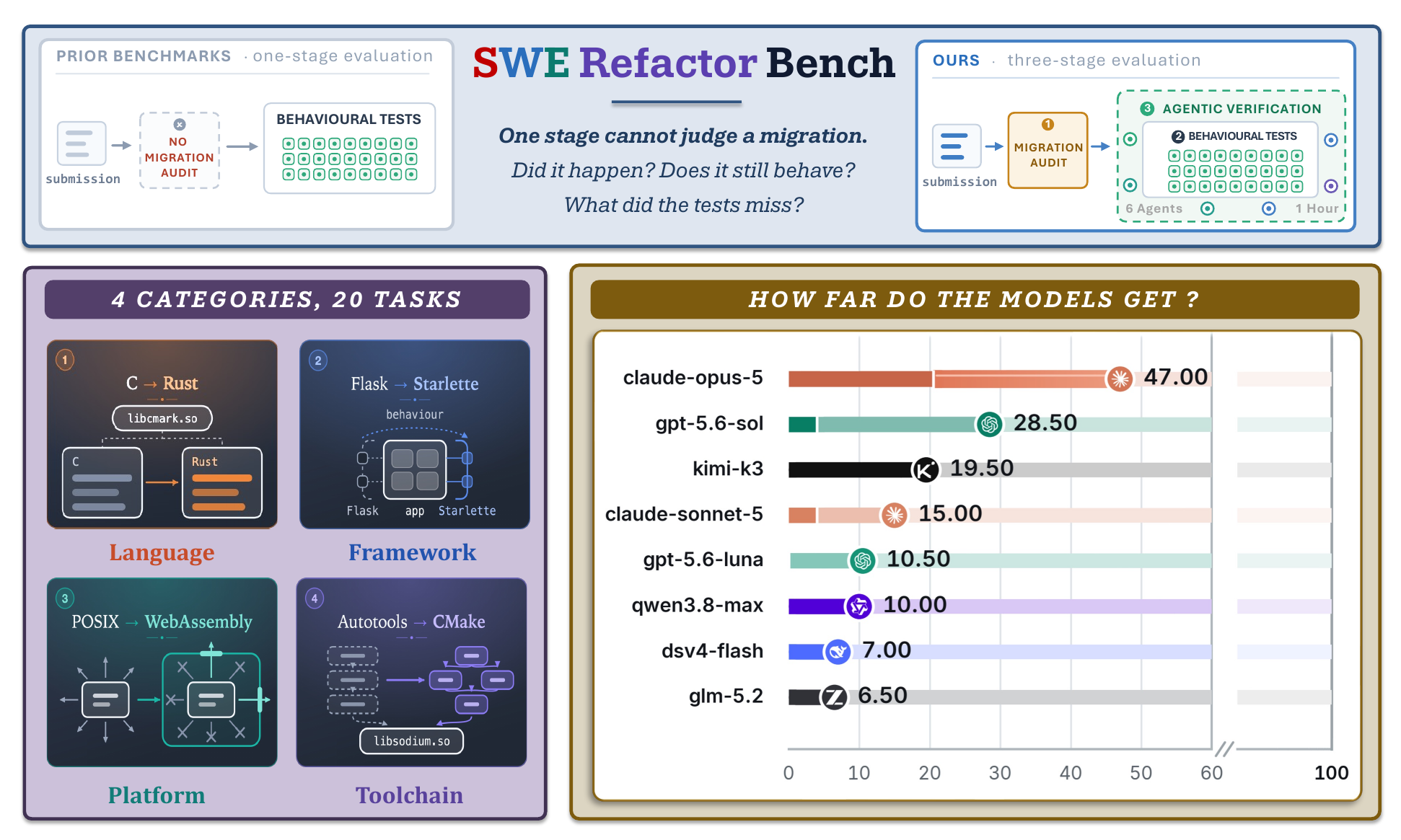}
  \caption{\textbf{\bench{} at a glance.} \textbf{Top:} how a migration gets scored. \textbf{Left,} prior
  benchmarks: one stage, a behavioural test suite. With no migration audit in front of it, a submission that
  never migrated anything still turns every check green, so the suite cannot tell a completed migration from an
  untouched repository. \textbf{Right,} \bench{}: three stages in series---\sI{} asks whether the migration
  happened at all, \sII{} is that same fixed suite, and \sIII{} sends six coding agents, one hour each, after
  the differences the suite was never written to catch. \textbf{Bottom left:} the 20 tasks span four kinds of technical
  debt---language, framework, platform, and build toolchain. \textbf{Bottom right:} composite scores of 8
  frontier models, each in its strongest configuration (out of 100; defined in~\eqref{eq:score}).}
  \label{fig:overview}
\end{figure}

% ── P1: scenario, opportunity, and capability gap ───────────────────────────
Modern software systems accumulate technical debt over decades, leaving long-lived repositories tied to
technology stacks that their teams no longer want to maintain~\citep{cunningham1992debt}. Replacing a language,
framework, platform, or build toolchain remains expensive and largely manual because engineers must coordinate
changes across source code, dependencies, interfaces, and build logic. Examples include rewriting a C library
in Rust and replacing Maven with Gradle across an entire repository. A successful migration must both replace
the old stack and preserve the system's behaviour. The original system provides a behavioural reference, so
correctness can be tested without a separate specification. Coding agents are increasingly reliable at
localized bug fixing~\citep{jimenez2024swebench,yang2024sweagent,wang2025openhands,zan2025multiswebench,zhang2025swebenchlive}.
This progress raises a harder question: can coding agents autonomously complete long-horizon, whole-repository
migrations while preserving behaviour? Reliable performance would extend coding agents from local edits to a
costly class of system maintenance.
Figure~\ref{fig:overview} contrasts the one-stage evaluation prior benchmarks apply with the three stages
\bench{} uses, and previews the benchmark's task coverage and model scores.

% ── P2: why existing benchmarks cannot answer the question ─────────────────
Existing benchmarks cannot establish whether coding agents can complete whole-repository migrations. They
evaluate repository-level code changes with fixed tests, including bug fixes, language-version upgrades, and
language migrations~\citep{jimenez2024swebench,zan2025multiswebench,zhang2025swebenchlive,migrationbench2025,wang2024repotransbench,khatry2025crustbench}.
These tests provide a clear red-to-green signal for bug fixing: at least one test fails before the patch and
passes after it. Whole-repository migration, however, starts from a repository whose tests already pass. A
correct migration and an untouched repository can therefore receive the same perfect test score. Fixed tests
measure behavioural correctness, but they cannot establish whether the migration actually occurred. This
creates an easy shortcut: agents can retain or copy the original implementation to make the tests pass. We call
this failure mode \emph{Blindness}: a behaviour-only evaluator awards full credit without establishing that the
target stack replaced the original implementation. Such an evaluator can detect whether known behaviour was
broken, but not whether the migration occurred.
This failure reflects a mismatch between what fixed tests observe and what migration requires. Fixed tests
compare observable behaviour, whereas migration completeness asks whether the target stack actually replaced
the original implementation. Adding more behavioural checks strengthens the evidence that behaviour was
preserved, but it cannot establish replacement: the untouched implementation passes those checks by
construction. Conversely, verifying replacement does not establish that behaviour survived. Evaluating
migration therefore requires separate evidence for both properties.

% ── P3: benchmark and evaluation protocol ───────────────────────────────────
We introduce \textbf{\bench{}}, which collects three forms of evidence through a three-stage protocol. \sI{}
is a hard gate that checks whether the target stack replaced the old implementation; any failed criterion
vetoes the submission. This gate prevents unchanged code and wrapper patches from earning credit. The \sII{}
stage measures behavioural correctness with $130{,}118$ fixed checks recorded from the original system and
requires every check to pass. \sIII{} tests behaviours that a fixed suite may
miss: six independent coding agents receive the original and migrated codebases
and one hour each to generate differential tests. These tests are generated after submission, vary across
verifier runs, and remain invisible to the migration agent. A verifier can reject a submission only with an
executable counterexample that passes on the original and fails on the migrated system. The three stages
prevent migration shortcuts, enforce known behavioural requirements, and probe differences not encoded in the
fixed suite.

% ── P4: task set, experimental setting, and core results ────────────────────
The benchmark comprises 20 whole-repository migrations of real open-source infrastructure, including SQLite,
zlib, libsodium, and GraphHopper. The tasks cover language (7), framework (7), platform (3), and build toolchain
(3) migrations and give agents 6 to 30 hours of autonomous work per task. Each task requires the repository to
use the target stack throughout while ensuring behavioural correctness. We evaluate 8 frontier models under
26 model--effort configurations, running each configuration once on all 20 tasks for 520 scored runs. Current
agents remain far from reliable whole-repository migration. The best configuration, \texttt{claude-opus-5} at
\textsf{xhigh} effort, scores
$47.0/100$; only 28 of 520 runs ($5.4\%$) pass all three stages, and 13 of the 20 tasks receive no accepted
solution. Migration completeness and behavioural correctness are distinct abilities, and agents miss them in
opposite directions: $30$ runs preserved behaviour by skipping the migration, and were stopped at \sI{}; $252$
completed the migration but broke behaviour, and were stopped at \sII{}. Agents also struggle to close the final correctness gap. Among
the 340 runs that pass \sI{}, $58\%$ reach $99\%$ of the fixed checks, yet only $26\%$ reach $100\%$. Agent
capability differs across migration categories: agents score $31.4$ on build toolchain rewrites but only $5.6$
on language rewrites. These results establish a capability gap on the 20 tasks in \bench{}; they do not rank
the intrinsic difficulty of all migration projects.

% ── P5: contributions ───────────────────────────────────────────────────────
In summary, we make the following contributions:
\begin{itemize}[leftmargin=1.4em,itemsep=2.5pt,topsep=3pt,parsep=0pt]
\item \textbf{Blindness and benchmark.} We identify \emph{Blindness}: behaviour-only evaluation can award full
credit without establishing that a migration occurred. And we introduce \bench{}, a benchmark of 20 long-horizon,
whole-repository migrations drawn from real open-source infrastructure and spanning language, framework,
platform, and build toolchain migrations.
\item \textbf{Three-stage adversarial evaluation.} We design a protocol that combines a hard migration audit
and $130{,}118$ fixed behavioural checks with agentic verification. During \sIII{}, six
independent coding agents actively search for hidden behavioural differences after submission and can reject a
migration only with an executable counterexample.
\item \textbf{Revealing agent capability gaps.} ``Getting the code right'' is not the same as ``getting the
migration done''---and even doing both is not enough. Agents miss the two conditions in opposite
directions---$30$ runs skip the migration, $252$ complete it and break behaviour---so neither stage can stand
in for the other; and among the 88 submissions that satisfy both under the fixed suite, agentic verifiers
break 60. Only 28 of 520 runs ($5.4\%$) survive all three stages. Current agents therefore rarely deliver migrations
that are complete, behaviour-preserving, and robust to agentic verification.
\end{itemize}

%% file: Sections_en/2_benchmark.tex
\section{\bench{}: Benchmark Design}
\label{sec:benchmark}

\bench{} measures one ability: given a working repository built on one technology stack, can an agent deliver
the same repository on another stack, with its observable behaviour intact and the old stack gone.

\subsection{Task Formulation}
\label{subsec:task_formulation}

A behaviour-preserving migration task is a tuple
\begin{equation}
  \tau = (R_A,\; \Sigma_A \rightarrow \Sigma_B,\; \mathcal{O},\; \mathcal{I},\; E,\; B).
  \label{eq:task}
\end{equation}
$R_A$---which we call \stA{}---is a real open-source repository, taken at one commit, in a buildable and
working condition. $\Sigma_A$ is the stack it is built on and $\Sigma_B$ the stack it must be built on
afterwards: a language, an application framework, a host platform, or a build toolchain. $\mathcal{O}$ is the
\emph{observable interface} of the artifact, that is, the surface available for inspection once the repository
has been built---process output and exit status, the symbols exported by an installed library, the manifest of
an installation tree, the responses of a served endpoint. $\mathcal{I}$ is the instruction given to the agent,
$E$ an offline image and $B$ a time budget.

The agent's \emph{input} is $(R_A, \mathcal{I}, E, B)$: a container holding the original repository, the
toolchains of both stacks, and an instruction saying what is to be done, with no network. Its \emph{output} is
the working tree $R_S$ at the end of the run. Writing $\mathcal{O}(R)$ for the observations produced by the
artifact that repository $R$ builds, $R_S$ solves $\tau$ exactly when both of the following hold:
\begin{align}
  &\Sigma_B \text{ builds the delivered artifact, and } \Sigma_A \text{ is absent from the repository and from the build closure,} \label{eq:migration}\\
  &\mathcal{O}(R_S) = \mathcal{O}(R_A). \label{eq:preservation}
\end{align}
We call \eqref{eq:migration} the \emph{migration condition} and \eqref{eq:preservation} the \emph{preservation
condition}. The first is a claim about the text of a repository and about what its build compiles; the second
is a claim about the behaviour of an artifact. No new functionality has to be designed anywhere in this work:
\stA{} already does everything \stB{} must do; the whole difficulty is to rebuild those behaviours exactly, on
a stack that expresses them in an entirely different way.

\paragraph{Why behavioural tests fail here.} A behavioural test suite $T$ is a finite set of observations drawn
from $\mathcal{O}$; the score it gives a repository $R$ is the fraction it passes, $\mathrm{rate}(R; T)$. Since
the preservation condition \eqref{eq:preservation} is defined \emph{relative to $R_A$} and $T \subseteq
\mathcal{O}$, we have $\mathrm{rate}(R_A; T) = 1$ for any $T$---by construction, and for every task in this
family. Setting $R_S = R_A$ (the \emph{empty diff}: the agent hands the repository back untouched) therefore
earns full marks on any behavioural suite while satisfying not one clause of the migration condition
\eqref{eq:migration}. This is exactly the \emph{blindness} of Section~\ref{sec:introduction}: the maximum of
the reward sits on a submission with zero work in it, and enlarging $T$ does not help, because every case the
migrated repository must pass is a case the original already passes.

\subsection{Where the Tasks Come From}
\label{subsec:composition}

\input{Tables_en/TabCatalog}

\paragraph{Pick the debt first, the repository second.} We did not choose repositories and then invent a change
for them; we went the other way: fix on a migration that a maintainer would call overdue, then look for a
project where that migration \emph{is} the whole job. Three admission requirements follow. First, \emph{the old
stack has to be load-bearing rather than incidental}, so that removing it reaches the design and not merely the
import statements---cmark's parser state machine, for instance, is hand-written C, and moving it to Rust means
redesigning ownership rather than adding a layer of FFI. Second, \emph{the observable interface has to be one
that something outside actually depends on}, so that ``behaviour preserved'' is a constraint with content
rather than a slogan---a C ABI, an HTTP API, an installation tree all have downstream consumers. Third,
\emph{there has to be a runnable reference}: \stA{} itself can be built and executed repeatedly, which is what
makes differential testing possible. We deliberately chose load-bearing infrastructure rather than exercises:
SQLite~\citep{gaffney2022sqlite,sw:sqlite}, zlib and the DEFLATE format it
implements~\citep{deutsch1996deflate,sw:zlib}, libsodium~\citep{bernstein2012nacl,sw:libsodium}, and
GraphHopper, whose routing is built on contraction hierarchies~\citep{geisberger2008ch,sw:graphhopper}, among
others.

\paragraph{Four kinds of technical debt.} A repository's stack comes down to four things: what language it is
written in, what framework it is organised around, what host it assumes, and what builds it. Each class of task
moves exactly one of them (Table~\ref{tab:catalog}). \emph{Language rewrites} (7 tasks) replace the
implementation language while keeping the shape of the artifact: the original language's implementation details
are deleted along with its source, yet the external behaviour of the artifact has to be reproduced exactly, and
the new language will not make the same choices by itself. \emph{Framework rewrites} (7) keep the language and
replace the dependency the application is organised around: most of what has to be reproduced is not code the
application wrote but what the framework did on its behalf---how errors are represented, how parameters are
coerced, how content is negotiated. \emph{Platform ports} (3) replace the host the code assumes: from POSIX to
\texttt{wasm32-wasi}~\citep{haas2017wasm}, process control, the filesystem namespace and memory mapping are
withdrawn, leaving only what the host explicitly grants. \emph{Build-toolchain rewrites} (3) move the part that
produces the artifact rather than the part that runs: what is compared is not only what the program does when
it runs but what it was packaged into.

\paragraph{What a task ships.} Once \eqref{eq:task} is instantiated, the agent receives: a \stA{} that builds
cleanly, a \stB{} declaration naming the target stack down to specific versions, an instruction stating the
requirements and how the build is invoked, an offline image---which \emph{also} carries \stA{}'s own
toolchain, since the original is its reference---and an artifact contract saying what will be collected from
the finished workspace. The rest it arranges itself: read the repository and the requirements, form a plan,
rewrite the code onto the target stack, get it to build into a real artifact, and finally check its own work
against \stA{}. On the evaluation side sit three things it never touches: the \sI{} criteria, written as prompt
questions about this repository; the \sII{} tests, whose expectations are recorded from a reference build and
run of \stA{} inside the evaluation image; and the further checks the six verifiers of \sIII{} perform beyond
those fixed tests. All three live in a separate image that is never mounted into the agent's container, so
those tests are not merely ``unread'' by it---they do not exist for it.

\subsection{The Three-Stage Protocol}
\label{subsec:protocol}

\begin{figure}[!htbp]
  \centering
  \includegraphics[width=\textwidth]{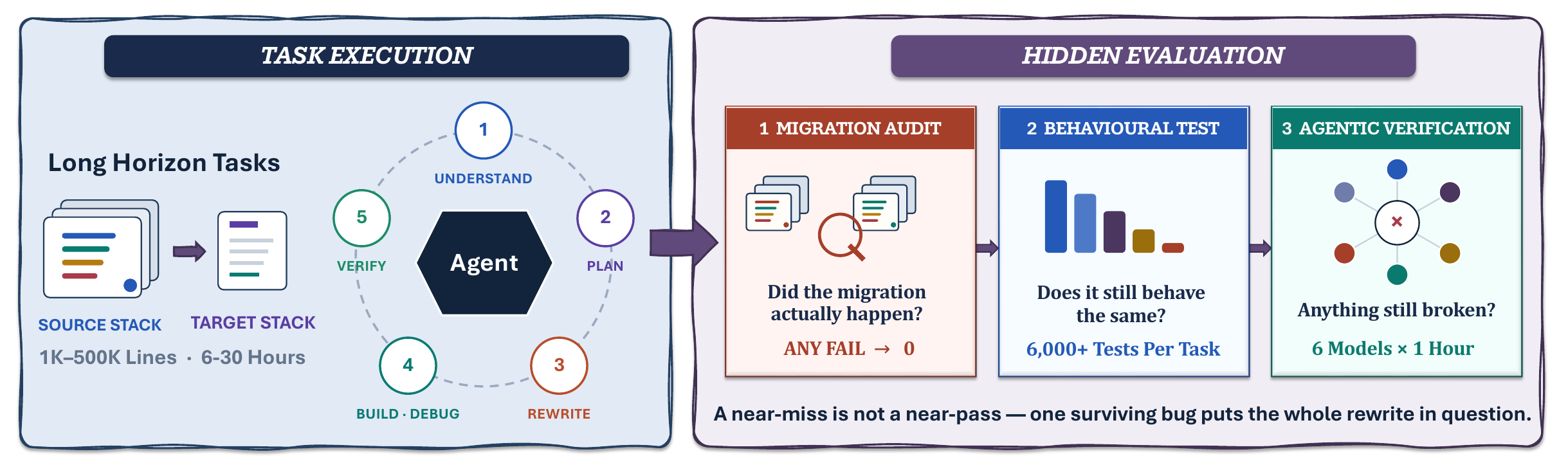}
  \caption{\textbf{On the left, what the agent sees}: a real repository on the source stack, an instruction, an
  offline image, and the process it works through on its own---read the repository and the requirements,
  rewrite onto the target stack, get it to build, then check itself against the original. \textbf{On the right,
  the three-stage evaluation it never touches}: \sI{} checks whether the migration actually happened, \sII{}
  checks whether the fixed behaviour is preserved exactly, and \sIII{} goes looking for whatever differences
  remain beyond the fixed tests.}
  \label{fig:process}
\end{figure}

Figure~\ref{fig:process} separates the agent's workspace and workflow from the hidden three-stage evaluation.
A submission is the working tree at the end of the agent's run; nothing it built is carried into evaluation,
and each stage rebuilds from the submitted source whatever it needs. The three stages are applied in order, and
each of them can end the run---except that we ran the fixed suite on every submission, vetoed or not, since
blindness cannot be counted otherwise; those runs still score zero.

\paragraph{Stage I: \sI{}.} It asks one thing: \emph{did the migration actually happen}---has the old stack
disappeared from the repository and from the build. The blindness of Section~\ref{sec:introduction} is exactly
why someone has to ask: a behavioural suite gives full marks to a repository handed back untouched, so ``was
the work done at all'' has to be checked separately. The criteria are written as prompt questions about this
repository, answered one by one by a model reading both source trees; every failing verdict must cite
re-checkable evidence, and each criterion is judged three times independently, with the majority taken.

\paragraph{Stage II: \sII{}.} It asks whether \emph{behaviour survived unchanged}. These tests were not written
out of thin air but built from \stA{}: the same calls are run against the original repository and the recorded
output becomes the expected answer, so the same inputs must afterwards produce the same outputs and the same
exit status. Across the suite there are $130{,}118$ checks, on average more than six thousand per task; a
single wrong check scores zero, and only a clean sweep opens Stage III.

\paragraph{Stage III: \sIII{}.} The first two stages check what we thought of in advance; this one checks
\emph{what we did not}. A fixed suite can only ask what its author happened to think of, and a submission that
reaches this point has answered all of it correctly. So we change the kind of checker---six \emph{coding
agents}, one hour each, each holding both source trees and looking for something the original repository does
that this submission does not. The six are divided up: five each probe one assigned direction (a C ABI, a set
of routes, an installation tree, and so on) and the sixth is unrestricted, so they complement one another in
coverage instead of re-searching the same ground. What they hand in cannot be a report, only an executable test
case: passing on \stA{} and failing on the submission, and it must first go green against the reference and
then reproduce three times, so that a broken test and a flaky test both fail to win. This is differential
testing~\citep{mckeeman1998differential} in an agentic form: a verifier may work the way a real tester
does---generate inputs against a stated property~\citep{claessen2000quickcheck}, relate one output to
another~\citep{chen2018metamorphic}, or shrink a difference until it is reportable~\citep{zeller2002delta}---but
what it says does not count; only a difference that runs does.

\paragraph{Scoring.} Let $g$ be the \sI{} verdict, $r_i$ the pass rate of module $i$ in \sII{}, and $s \in
\{0,\dots,6\}$ the number of verifiers that failed to produce a counterexample. A submission scores
\begin{equation}
  S(\tau) \;=\; \underbrace{\mathbf{1}\bigl[g = \textsf{pass}\bigr]}_{\text{Stage I: veto}}
  \cdot \underbrace{\mathbf{1}\bigl[r_i = 1 \;\; \forall i\bigr]}_{\text{Stage II: all or nothing}}
  \cdot \Bigl(\, 0.4 \;+\; \underbrace{0.6 \cdot \tfrac{s}{6}}_{\text{Stage III: per verifier}} \Bigr)
  \;\in\; \{0\} \cup [0.4,\, 1].
  \label{eq:score}
\end{equation}
Every factor in the formula corresponds to one condition from Section~\ref{subsec:task_formulation}. Stage I
\emph{multiplies rather than adds}, because without the migration the task was not done at all, and no amount
of behavioural correctness should buy points for it. Stage II is \emph{all or nothing}, because what this
family of tasks asks is whether this is a \emph{drop-in} replacement, and a library that is wrong once in a
thousand calls is not: behind the failing test stands a downstream consumer, and it will not be spared because
the other $99.99\%$ of the behaviour is right (Section~\ref{subsec:lastmile} gives concrete examples). Stage
III carries $0.6$ because what it looks at is a residue that no fixed suite can see, and so deserves the larger
share; it is scored linearly rather than all-or-nothing because \emph{not broken} is not the same as \emph{no
difference}: six verifiers finding nothing is more credible than one verifier finding nothing, but this is
still a matter of how strong the evidence is, not a demonstration of equivalence, and a linear score records
precisely that degree.

%% file: Tables_en/TabCatalog.tex
\begin{table}[!t]
\centering
\footnotesize
\setlength{\tabcolsep}{4pt}
\renewcommand{\arraystretch}{1.1}
\begin{tabular}{@{}p{0.135\linewidth}p{0.275\linewidth}rrrrrr@{}}
\toprule
\textbf{Debt class} & \textbf{Representative migrations} & \textbf{Tasks} &
\textbf{Source LoC} & \textbf{Budget} & \textbf{Criteria} & \textbf{Modules} & \textbf{Checks} \\
& & & & \textbf{(h)} & \textbf{(Stage I)} & \textbf{(Stage II)} & \textbf{(Stage II)} \\
\midrule
Language      & C\,$\to$\,Rust, C\,$\to$\,Java, Go\,$\to$\,Zig
              & 7 & 4.4\,k--39.8\,k & 12--30 & 60 & 117 & 59{,}771 \\
Framework     & Flask\,$\to$\,Starlette, Gin\,$\to$\,chi, Vue\,$\to$\,React
              & 7 & 0.8\,k--94.8\,k & 6--16  & 39 & 79  & 55{,}852 \\
Platform      & POSIX\,$\to$\,\texttt{wasm32-wasi}, Node\,$\to$\,V8 realm
              & 3 & 16.0\,k--358.0\,k & 6--10 & 19 & 31 & 6{,}725 \\
Build toolchain & Autotools\,$\to$\,CMake, Maven\,$\to$\,Gradle
              & 3 & 19.0\,k--78.8\,k & 6      & 18 & 37 & 7{,}770 \\
\midrule
\textbf{Total} & \textbf{20 upstream projects, 20 tasks} & \textbf{20} &
\textbf{867{,}062} & \textbf{262} & \textbf{136} & \textbf{264} & \textbf{130{,}118} \\
\bottomrule
\end{tabular}
\caption{\textbf{Composition of the \bench{} task set.} Every task requires a real open-source repository to be
migrated in its entirety onto another stack, with its interface and observable behaviour unchanged.
\emph{Budget} is the time limit given to the agent; \emph{criteria} are the Stage I migration questions,
\emph{modules} are the independent test modules of Stage II, and \emph{checks} are the behavioural cases
those modules hold.}
\label{tab:catalog}
\end{table}

%% file: Sections_en/3_experiments.tex
\section{Experiments}
\label{sec:experiments}

This section first states the setup and the metrics (Section~\ref{subsec:setup}), then answers two questions:
how far today's agents get on whole-repository behaviour-preserving migration
(Sections~\ref{subsec:main_results}--\ref{subsec:analysis}), and whether the evaluation itself holds up
(Section~\ref{subsec:validation}).

\subsection{Setup and Metrics}
\label{subsec:setup}

We evaluate eight frontier models: \texttt{claude-opus-5} and \texttt{claude-sonnet-5};
\texttt{gpt-5.6-luna} and \texttt{gpt-5.6-sol}; \texttt{kimi-k3}~\citep{kimik2};
\texttt{qwen3.8-max}~\citep{qwen3}; \texttt{dsv4-flash}~\citep{deepseekv3}; and
\texttt{glm-5.2}~\citep{glm45}. The two GPT-series models use Codex as their harness; the other six use Claude
Code. Every model ran all 20 tasks, for a
total of $8$ models, $26$ configurations and $520$ scored runs. A run is one independent attempt by one model
at one task: a fresh container is created from that task's image, the instruction is handed to the agent, and
it works on its own within the time budget the task declares, with no network beyond the model endpoint; when
the budget runs out or it stops of its own accord, the working tree is collected as a submission and sent into
the three-stage evaluation.

\paragraph{Six metrics, each answering a different question.} Alongside the composite score of
\eqref{eq:score} we report counts, because a count states \emph{what happened to} a submission, whereas a
score only states it after weighting. The following definitions are used throughout the paper.
\begin{itemize}[leftmargin=1.4em,itemsep=1.5pt,topsep=3pt,parsep=0pt]
\small
\item \textbf{Migrated} (Stage I passed): runs in which every migration criterion was judged to pass by
majority.
\item \textbf{All tests pass} (Stage II perfect): runs that passed every fixed check in \sII{}.
\item \textbf{Accepted}: runs that migrated, passed every fixed check, and survived all six verifiers.
\item \textbf{Broken}: the \emph{share} of runs that migrated and passed every fixed check but were then
defeated by at least one verifier.
\item \textbf{Blindness}: runs that passed every fixed check yet were vetoed at Stage I. A behavioural
instrument gives full marks to a repository that was never migrated---this count corresponds directly to the
\emph{blindness} phenomenon of Section~\ref{sec:introduction}.
\item \textbf{Score}: the mean over runs of the score from~\eqref{eq:score}, out of $100$.
\end{itemize}

\subsection{Overall Performance}
\label{subsec:main_results}

\input{Tables_en/TabMain}

% \begin{figure}[!htbp]
% \centering
% \includegraphics[width=\textwidth]{Figures/stages.pdf}
% \caption{\textbf{(a)} The same $520$ runs, judged two ways side by side. The upper bar is the three-stage
% protocol: five mutually exclusive and exhaustive terminal states, ordered so that the runs passing every test
% form a prefix on the left. The lower bar is the verdict of the \emph{fixed suite alone} on the same runs: at
% full marks it is one indivisible block, because once every test passes it has nothing left to sort by.
% \textbf{(b)} The same five states per model, normalised and ordered by composite score.}
% \label{fig:stages}
% \end{figure}

\paragraph{Leaderboard.} Table~\ref{tab:score} lists all $26$ configurations with their behavioural pass rate,
score and cost, with each model's best-scoring row shaded. Comparing strongest configurations,
\texttt{claude-opus-5} leads on acceptances and score: 5 acceptances in $20$ runs and a score of
$47.0$; behind it are \texttt{gpt-5.6-sol} at $28.5$, \texttt{kimi-k3} at $19.5$ and \texttt{claude-sonnet-5}
at $15.0$. Across the eight models in their strongest configurations, $160$ runs yield only $14$ acceptances.

The rows worth studying are the three with no acceptances at all. \texttt{gpt-5.6-luna}, \texttt{dsv4-flash}
and \texttt{glm-5.2} never got one, yet each produced runs that passed every fixed check---$4$, $3$, and $3$,
respectively, scattered across the \emph{blindness} and \emph{counterexample found} columns of
Table~\ref{tab:main}---and those runs then fell at Stage I or Stage III. Judged by the fixed checks alone,
each of these three models would appear on the leaderboard with a few ``perfect scores''; under the three-stage
protocol they solved nothing.

\paragraph{The funnel is narrow, and it narrows for three different reasons.} Table~\ref{tab:main} decomposes
each row's $20$ runs into five mutually exclusive outcomes. Of the $520$ scored runs, $340$ ($65.4\%$) passed
Stage I and $118$ ($22.7\%$) passed every fixed check, but only $88$ did both and thereby reached Stage
III, of which $28$ ($5.4\%$) survived all six verifiers; 13 of the 20 tasks were never solved by any model. The
mean score over all $520$ runs is only $13.44/100$, but over the $88$ that reached Stage III it is $79.43$---so
the difficulty is in getting to Stage III at all. Moreover the three gates stop \emph{different} submissions;
no group is filtered out twice: stages that overlap are redundant, whereas stages that are disjoint are
measuring three different things. With the fixed suite as the only instrument, $118$ submissions would tie for
first place, among them submissions that never migrated anything and submissions that a verifier breaks within
the hour.

Outcomes also vary by migration category (Table~\ref{tab:category}). Build toolchain rewrites have the highest
Stage I pass rate ($80.8\%$) and mean score ($31.4$), whereas framework rewrites contribute $14$ of the $28$
accepted runs.

\subsection{Analysis of Agent Behaviour}
\label{subsec:analysis}

\subsubsection{``Getting the code right'' and ``getting the migration done'' are two different abilities}
\label{subsec:shortcut}

\begin{figure}[!htbp]
\centering
\includegraphics[width=\textwidth]{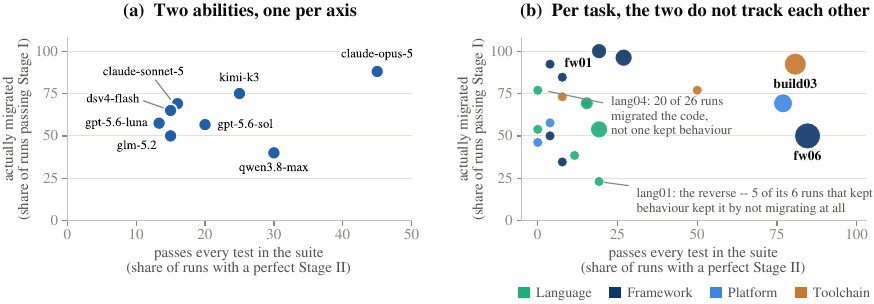}
\caption{\textbf{(a)} Where each model sits on two axes, both of them shares of the same runs: horizontally
``does it still behave like the original'' (a perfect Stage II), vertically ``was the migration done at all''
(Stage I). \textbf{(b)} The same two axes, per task, with marker area the number of accepted submissions. A
migration has to satisfy both conditions, but they are satisfied by \emph{different} runs, and most runs
manage at most one.}
\label{fig:shortcut}
\end{figure}

Put the two axes together and the failures fall into two groups
(Figure~\ref{fig:shortcut}a). One is the runs that never migrated and kept every fixed check: a submission
that barely migrates anything passes any behavioural suite by construction, and $30$ runs did exactly that,
spread over $7$ of the $8$ models. Stage II gives all $30$ full marks; only Stage I stops them. The other is
eight times as large: $252$ runs completed the migration and broke behaviour doing it, and only Stage II stops
those. Neither stage can stand in for the other, then: Stage II alone would reward doing nothing, Stage I
alone would reward doing damage.

\begin{figure}[!htbp]
\centering
\includegraphics[width=\textwidth]{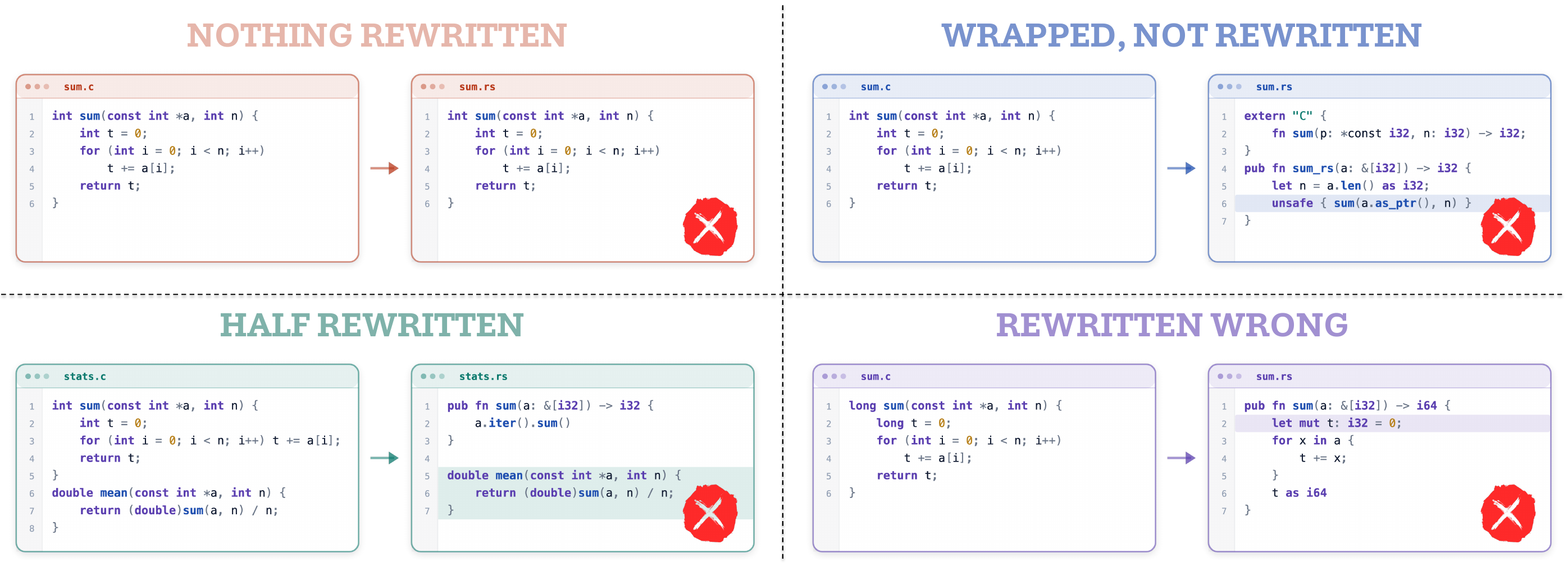}
\caption{\textbf{The four kinds of submission Stage I has to tell apart.} Top left, \emph{nothing rewritten}:
the new file is the C copied over verbatim, only the suffix changed. Top right, \emph{wrapped}: the Rust side
is an \texttt{extern "C"} forwarding shim and the original C still does the work. Bottom left, \emph{half
done}: some functions are genuinely rewritten while others still call back into the old implementation. Bottom
right, \emph{rewritten wrongly}: really rewritten, but the behaviour changed. The first two pass every
behavioural test, and only an instrument that reads mechanism can reject them.}
\label{fig:errorcases}
\end{figure}

Per task, the two axes diverge just as clearly (Figure~\ref{fig:shortcut}b), and in both directions. On
\texttt{lang04} (acorn, JavaScript $\to$ Rust), $20$ of $26$ runs passed Stage I, yet \emph{not one} passed
every fixed check. Twenty runs completed the migration, but all $26$ failed at least one fixed check.
\texttt{lang01} (cmark, C $\to$ Rust) is the mirror image: only $6$ runs passed Stage I, yet $5$ passed every fixed check---and
\emph{all five} of those are blindness. Each cleared seven of the eight criteria and failed only the one
asking whether the Rust is the implementation: it reproduces the original's control flow statement for
statement, a transliteration---ownership was never redesigned, the original's manual memory management simply
moved into Rust, and the safety this migration exists to buy was not bought. No behavioural test can express
that, because behaviour was never where the problem was. Figure~\ref{fig:errorcases} shows the four
typical shapes of such submissions; the first two (handed back as-is, wrapped in a shim) are defects no
behavioural test can express.

% \begin{figure}[!htbp]
% \centering
% \includegraphics[width=\textwidth]{Figures/blindness.pdf}
% \caption{\textbf{(a)} Pass rate of the fixed suite on each task when the \emph{original repository is handed
% back untouched} (not one character changed). \textbf{(b)} The same instrument ranking the eight real models
% together with that empty diff. Measurement convention: the denominator is the number of checks for which that
% evaluation actually recorded a verdict (pass or fail); ``not one failure'' is $100$. One task's control record
% was written under another account and unreadable at analysis time, hence 19 tasks.}
% \label{fig:blindness}
% \end{figure}

\subsubsection{The last 1\%: agents cannot deliver a perfect migration}
\label{subsec:lastmile}

\begin{figure}[!htbp]
\centering
\includegraphics[width=\textwidth]{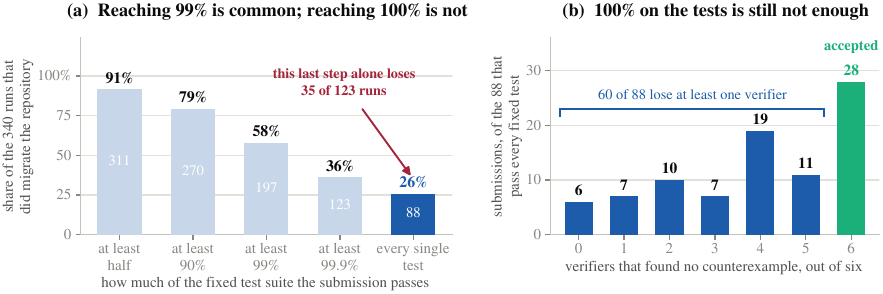}
\caption{\textbf{(a)} Among the $340$ runs that \emph{did} complete the migration, the share reaching each
level on the fixed suite. \textbf{(b)} Among the $88$ submissions that passed every fixed check, how many
of the six independent verifiers (one hour each) failed to find a behavioural difference. Only the rightmost
bar is an accepted migration.}
\label{fig:lastmile}
\end{figure}

For a repository migration, any failing unit test is a serious risk: behind that failing test stands a real
downstream consumer, and it will not be spared because the other $99.99\%$ of the behaviour is right. And this
is precisely the step agents cannot get past. Among the $340$ runs that passed Stage I
(Figure~\ref{fig:lastmile}a), $91\%$ get the fixed suite past half, $58\%$ reach $99\%$ and $36\%$ reach
$99.9\%$---but only $26\%$ make no error at all. That final step alone eliminates $35$ of the $123$ runs that
had already reached $99.9\%$; across the campaign $140$ runs land in $[99\%, 100\%)$, missing a median of
$12.5$ checks, and $18$ of them miss exactly one.

Those $18$ are not scattered at random; they concentrate on a handful of specific checks. On \texttt{fw03}
(conduit, Vue $\to$ React), four different models all ended at $21768/21769$, failing the same one: the
original uses hash routing, so visiting \texttt{/} settles at \texttt{/\#/} while the React version stays at
\texttt{/}, which breaks every bookmark and every shared link to the site. On \texttt{build03} (PyCryptodome,
setuptools $\to$ Meson), five different models all ended at $380/381$: the \texttt{METADATA} long description
in the built wheel is $0$ characters, so once the package is published its PyPI project page is blank. Not one
of these checks can be called nitpicking: each corresponds to a regression that would cause trouble in
production, and the original repository passes \emph{all} of them. In other words, this is not a problem with
the test suite; it is a problem with the migration.

Making no test error is still not enough. Among the $88$ submissions that did not miss a single fixed check,
only $28$ survived all six verifiers; the other $60$ ($68.2\%$) had a counterexample found against them within
the hour (Figure~\ref{fig:lastmile}b), with the average submission holding off only $3.94$ of the $6$
verifiers. In other words, an instrument that looks only at behaviour would have accepted all $88$ of these
submissions, and Stage III took back two thirds of them. When a break comes, it comes quickly: the median time
to a counterexample is $17.0$ minutes, against $32.8$ minutes for a survival.

\subsubsection{Agent capability differs across migration categories}
\label{subsec:category}

\input{Tables_en/TabStageCategory}

Agents exhibit distinct capability profiles across the four migration categories (Table~\ref{tab:stagecat}).
Their overall scores rank build toolchain rewrites first ($31.4$), followed by platform ports ($17.2$),
framework rewrites ($12.0$), and language rewrites ($5.6$). However, \keyclaim{agents do not perform best on
one category throughout the pipeline}. They achieve their highest Stage I and Stage II pass rates on build
toolchain rewrites ($80.8\%$ and $54.0\%$), but their lowest Stage III survival rate on the same category
($17.6\%$). On framework rewrites, agents achieve only $18.9\%$ at Stage II but their highest Stage III
survival rate, $56.0\%$.

\keyclaim{Agents encounter different bottlenecks across migration categories.} Their lowest conditional pass
rate occurs at Stage III for build toolchain rewrites and platform ports ($17.6\%$ and $23.5\%$), but at Stage
II for framework and language rewrites ($18.9\%$ and $12.0\%$). These profiles align with the repository
surface that agents must modify. In build toolchain rewrites and platform ports, agents mainly change how the
repository is built or which host it runs on while leaving the product code largely unchanged. Their
submissions pass Stage II relatively often, but Stage III exposes behavioural differences outside the fixed
suite. In framework and language rewrites, agents modify the product code itself, and the fixed suite
filters their submissions at Stage II before most reach the verifiers. Language rewrites show the sharpest
early attrition: only $12$ of $182$ runs reach Stage III.

\subsection{Analysis of Benchmark Validity}
\label{subsec:validation}

The premise of \bench{} is a claim about an \emph{instrument} rather than about models, so this section
examines the instrument itself: whether the Stage I judge is stable (Section~\ref{subsec:judge_ablation}), what
the Stage II and Stage III tests each ask (Section~\ref{subsec:hidden}), and whether all six verifiers are
needed (Section~\ref{subsec:adversary}).

\subsubsection{Stage I Decisions Agree Across Models and Humans}
\label{subsec:judge_ablation}

% \begin{figure}[!htbp]
% \centering
% \includegraphics[width=\textwidth]{Figures/judge.pdf}
% \caption{\textbf{Is asking a model ``was this migration completed?'' good enough.} \textbf{(a)} On the $126$
% valid verdicts, each judge family is plotted by ``how many unfinished submissions it caught'' (sensitivity)
% against ``how many accepted submissions it correctly let through'' (specificity). The dashed anti-diagonal is
% the chance line $se + sp = 1$, and the shaded triangle below it is worse than answering without reading the
% code. The hollow ring is all $126$ verdicts pooled. \textbf{(b)} On the $80$ cells whose material really does
% contain a defect that this protocol scores, where the judge's verdict and its stated reason fall.}
% \label{fig:judge}
% \end{figure}

\emph{Does the same judge model read the same tree differently three times?} The judge is
\texttt{gpt-5.6-sol}; every criterion is judged by three independent samples of it and the majority is taken. Of $3{,}536$ criterion verdicts, $3{,}405$ ($96.3\%$) had
all three samples agree, and only $131$ split $2{:}1$. That looks like very little disagreement, but Stage I
requires \emph{every} criterion to pass, so a single disagreement on a single criterion can decide the fate of
a whole run: of the $340$ runs that passed Stage I, $35$ ($10.3\%$) had at least one criterion that passed only
$2{:}1$---on that criterion, one sample argued for zero. Which means that without majority voting, letting any
single dissenting sample count, the number passing Stage I would be $305$ rather than $340$. Majority over
three samples is load-bearing here, not ceremonial.

\input{Tables_en/TabHumanEval}

\emph{Would a human judge differently?} We asked two software-engineering researchers not involved in this work
to decide independently, without seeing the judge model's verdicts, whether ``this is a genuine migration'' for
all $156$ runs of $6$ tasks spanning the four migration classes, three tasks each, working from the same
criteria the judge saw. Judge and human agree $89.7\%$ of the time ($140/156$, $\kappa = 0.795$), and the
direction of the disagreements matters more than their number: of the $16$ disagreements, $14$ are the judge
being too strict---the human considers the migration genuine and the judge scored zero---and only $2$ go the
other way; Table~\ref{tab:humaneval} breaks this down by task.
Following all $16$ through the later stages changes almost nothing about the final verdict: of the
$14$ too-strict cases, $12$ would have failed the full fixed suite at Stage II anyway; of the $2$ too-lenient
cases, $1$ likewise; and the remaining one reached Stage III, where four of the six agentic verifiers each
constructed a counterexample against it, so it was not accepted either. That is, the Stage I error let no
submission a human thought was disguised into the accepted set; what it may have undercounted is at most the
$2$ runs that passed every behavioural check and were zeroed by Stage I alone. The judge's error has a
direction, and that direction makes points harder to earn, not easier.

\emph{Does the judge spare its own family?} It is itself one of the evaluated systems, so leniency towards
submissions from the same family would quietly inflate them. That does not appear: \texttt{gpt-5.6-sol} passed
$57.1\%$ of the $240$ submissions written by a GPT-series model against $72.5\%$ of the other $280$. Most of
that gap is submission quality rather than provenance.

\subsubsection{Agentic Verification Finds Failures Beyond Fixed Tests}
\label{subsec:hidden}

Stage II and Stage III test two different parts of the same thing, and the clearest way to see it is to put
two sets of cases side by side.

\emph{What Stage II asks is what the task author could think of in advance.} The check from
Section~\ref{subsec:lastmile}---whether visiting \texttt{/} settles at \texttt{/\#/}---is an observation you
get simply by running the original once, which is why it sits in the fixed suite as one of $21{,}769$ checks.
Coverage of that kind is already wide: \texttt{fw03} alone carries more than twenty-one thousand checks.

\emph{What Stage III asks cannot be anticipated.} On \texttt{fw04} (ChartMuseum, Gin $\to$ chi), \texttt{POST
/api/charts} has to dispatch between a multipart-form upload handler and a raw-body upload handler; the
original truncates \texttt{Content-Type} at the first space \emph{or} semicolon, while the migrated version
truncates only at the semicolon, so a request with one extra space before the boundary reaches a different
handler on each side---an implementation detail of Gin whose existence the task author did not know of when
writing the suite. Others of the same kind: \texttt{lang05} (go-yaml, Go $\to$ Zig), where go-yaml, by way of
Go's \texttt{time.Parse}, also accepts a comma as the decimal separator, so
\texttt{2001-12-14T21:59:43,10Z} is a \texttt{!!timestamp} on the original and a \texttt{!!str} in the Zig
version; and \texttt{pf01} (SQLite, POSIX $\to$ WASI), where after checkpointing a WAL database and switching
it back to DELETE mode, native SQLite removes both the \texttt{-wal} and the \texttt{-shm} file while the WASI
version removes only \texttt{-wal}.

What the two kinds have in common is that each is decisive---if this one check does not pass, the migration is
not complete; the only difference is that one can be fixed in advance as a test and the other cannot. That is
exactly why both stages are indispensable, and exactly why Stage III has to hand in an \emph{executable}
program rather than a verdict: only an executable counterexample turns ``a problem nobody thought of'' into a
fact that can be re-checked.

\subsubsection{Six Verifiers: Strength and Diversity Both Matter}
\label{subsec:adversary}

\begin{figure}[!htbp]
\centering
\includegraphics[width=\textwidth]{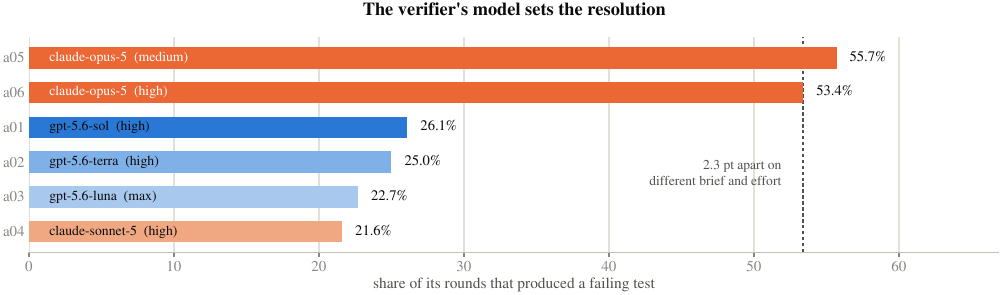}
\caption{\textbf{Break rate of each of the six verifiers}: the share of its rounds in which it handed in an
executable counterexample that passes on the original and fails on the submission, coloured by the model
behind it and annotated with its reasoning-effort setting. The dashed line marks the lower of the two
\texttt{claude-opus-5} verifiers.}
\label{fig:adversary}
\end{figure}

\paragraph{Finding 1: a strong coding agent as the verifier matters an order of magnitude more than its
configuration.} The six verifiers face the \emph{same} $88$ submissions and each runs $88$ rounds, so task
difficulty is identical for them; all that differs is which direction each was assigned. The break rates,
however, are far apart (Figure~\ref{fig:adversary}): the two held by \texttt{claude-opus-5} are $55.7\%$ and
$53.4\%$, the other four between $21.6\%$ and $26.1\%$. That gap cannot be attributed to opus happening to draw
the easy directions, for two reasons. First, of the two opus verifiers, one probes an assigned direction and
the other is unrestricted, so their prompts and effort settings \emph{both} differ, and yet their break rates
differ by only $2.3$ percentage points. Second, the \emph{unrestricted} one was given no direction at all---it
was free to search the same ground as the other four---and it still broke $53.4\%$. Changing prompt and effort
within one model moves the break rate by about two points; changing the model moves it by thirty. This also
means the reported numbers are a property of this panel as much as of the submissions: retire the two
strongest and the remaining four would accept $46$ submissions instead of $28$. An accepted submission is
therefore not a migration proven correct but a migration that survived the strongest adversaries we could
field---and as models improve, this same fixed set of 20 tasks will be scored more strictly.

Nor are the six redundant. Counting \emph{exclusive} breaks---where this verifier alone found a counterexample
and the other five did not---the two opus verifiers have $5$ and $2$, and the four weaker ones $4$ between
them. Even the one with the lowest break rate rejected a submission that the other five let through: they
differ in strength, but they are complementary in \emph{coverage}.

\paragraph{Finding 2: models do not spare their own family's work.} This matters, because
\texttt{claude-opus-5} both leads the leaderboard and holds two of the verifiers. Pooled, the effect runs
\emph{opposite} to collusion: verifiers broke $33.9\%$ of the submissions written by a model of their own
family and $34.3\%$ of everyone else's. The apparent counter-evidence is that the opus verifiers break $40.8\%$
of opus-authored submissions against $65.0\%$ of everyone else's; but that fails its own control---the four
non-opus verifiers on the same submissions are at $17.1\%$ against $29.1\%$, a drop \emph{by the same factor}.
What separates the two is submission quality, not collusion: submissions written by opus are simply harder to
break, for everyone.

%% file: Tables_en/TabMain.tex
% Generated from the raw scoring tables under Figures/src. Shading encodes only
% the magnitude of a value within its column; the arrow in the header says
% whether more or less is better for that column.
\newcommand{\g}[2]{\cellcolor{brandprimary!#1}#2}
\newcommand{\bd}[1]{\cellcolor{brandbg}#1}
\newcommand{\up}{\raisebox{0.15ex}{\scriptsize$\uparrow$}}
\newcommand{\dn}{\raisebox{0.15ex}{\scriptsize$\downarrow$}}
\newcommand{\modelicon}[1]{\raisebox{-0.12ex}{\includegraphics[height=1.05em]{assets/model_icons/#1.png}}}

\definecolor{sIa}{HTML}{C0503D}
\definecolor{sIb}{HTML}{A03A2B}
\definecolor{sII}{HTML}{E0912F}
\definecolor{sIII}{HTML}{C9A227}
\definecolor{sOK}{HTML}{3F8F55}
\newcommand{\captionchip}[2]{{\setlength{\fboxsep}{0pt}\colorbox{#1}{\strut\,#2\,}}}

\begin{table}[!t]
\centering
\footnotesize
\setlength{\tabcolsep}{2.4pt}
\renewcommand{\arraystretch}{1.15}
\caption{\textbf{The three-stage funnel: where each of the $520$ runs stops.} Each row aggregates $20$ runs,
one per task. Values are numbers of runs; the five outcome columns are mutually exclusive and exhaustive, so
they sum to $20$ across a row.
\emph{Harness} identifies the coding-agent client: GPT-series models use Codex, and all others use Claude Code.
Colour marks the stage at which a run stops: \captionchip{sIa!45}{\sI{}},
\captionchip{sII!45}{\sII{}}, \captionchip{sIII!45}{\sIII{}},
\captionchip{sOK!45}{all three passed}; within a column, darker means more runs in that cell.
\emph{Blindness} is a special kind of Stage I failure: all fixed checks pass, yet the repository was never
migrated.}
\label{tab:main}
\begin{adjustbox}{max width=\textwidth}
\begin{tabular}{@{}l l l @{\hspace{4pt}} >{\centering\arraybackslash}m{2.1cm} >{\centering\arraybackslash}m{2.1cm} @{\hspace{4pt}} c @{\hspace{4pt}} c @{\hspace{4pt}} c@{}}
\toprule
\multicolumn{1}{c}{\multirow{2}{*}{\textbf{Model}}}
& \multicolumn{1}{c}{\multirow{2}{*}{\textbf{Harness}}}
& \multicolumn{1}{c}{\multirow{2}{*}{\textbf{Effort}}}
& \multicolumn{2}{c}{\textbf{Stopped at Stage I: not migrated}} & \multicolumn{1}{c}{\textbf{Stopped at Stage II}}
& \multicolumn{1}{c}{\textbf{Stopped at Stage III}} & \multicolumn{1}{c}{\textbf{Passed}}\\
\cmidrule(lr){4-5}\cmidrule(lr){6-6}\cmidrule(lr){7-7}\cmidrule(lr){8-8}
& & & {\scriptsize\shortstack{fixed checks\\also fail}}
& {\scriptsize\shortstack{fixed checks pass\\(Blindness)}}
& \multicolumn{1}{c}{\scriptsize\shortstack{fixed checks\\fail}}
& \multicolumn{1}{c}{\scriptsize\shortstack{counterexample\\found}}
& \multicolumn{1}{c}{\raisebox{1.25ex}{\scriptsize accepted}}\\
\midrule
\modelicon{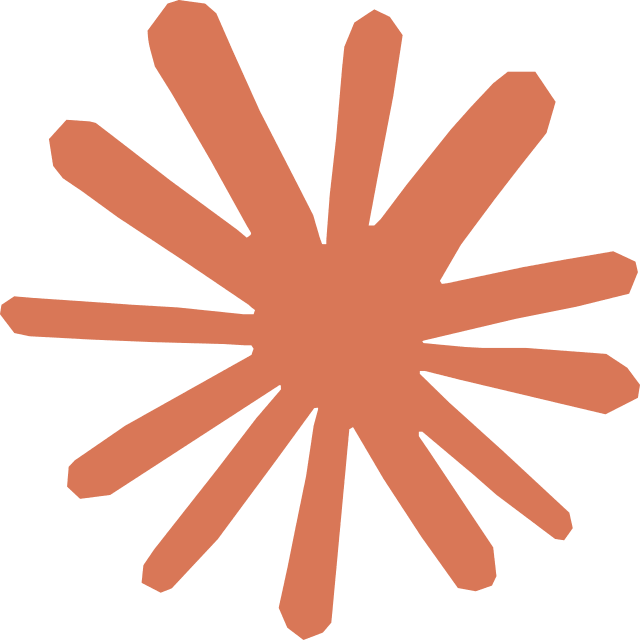}\hspace{0.35em}Claude Opus 5 & Claude Code & \textsf{low} & \cellcolor{sIa!22}2 & \cellcolor{sIb!6}\textcolor{black!55}{0} & \cellcolor{sII!68}13 & \cellcolor{sIII!43}3 & \cellcolor{sOK!37}\textbf{2} \\
\modelicon{claude}\hspace{0.35em}Claude Opus 5 & Claude Code & \textsf{medium} & \cellcolor{sIa!18}1 & \cellcolor{sIb!43}2 & \cellcolor{sII!55}10 & \cellcolor{sIII!62}5 & \cellcolor{sOK!37}\textbf{2} \\
\modelicon{claude}\hspace{0.35em}Claude Opus 5 & Claude Code & \textsf{high} & \cellcolor{sIa!6}\textcolor{black!55}{0} & \cellcolor{sIb!43}2 & \cellcolor{sII!55}10 & \cellcolor{sIII!53}4 & \cellcolor{sOK!60}\textbf{4} \\
\modelicon{claude}\hspace{0.35em}Claude Opus 5 & Claude Code & \textsf{xhigh} & \cellcolor{sIa!6}\textcolor{black!55}{0} & \cellcolor{sIb!28}1 & \cellcolor{sII!47}8 & \cellcolor{sIII!72}6 & \cellcolor{sOK!72}\textbf{5} \\
\modelicon{claude}\hspace{0.35em}Claude Opus 5 & Claude Code & \textsf{max} & \cellcolor{sIa!22}2 & \cellcolor{sIb!43}2 & \cellcolor{sII!51}9 & \cellcolor{sIII!53}4 & \cellcolor{sOK!49}\textbf{3} \\
\addlinespace[2pt]
\modelicon{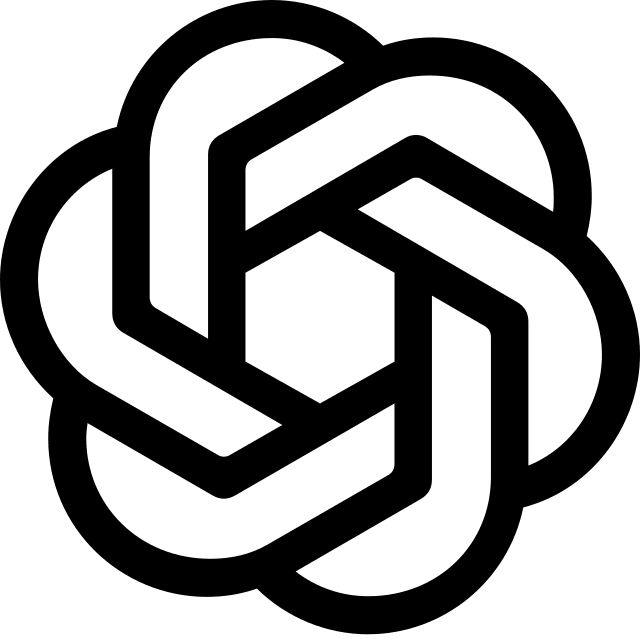}\hspace{0.35em}GPT-5.6 Sol & Codex & \textsf{none} & \cellcolor{sIa!60}11 & \cellcolor{sIb!6}\textcolor{black!55}{0} & \cellcolor{sII!47}8 & \cellcolor{sIII!24}1 & \cellcolor{sOK!6}\textcolor{black!55}{0} \\
\modelicon{openai}\hspace{0.35em}GPT-5.6 Sol & Codex & \textsf{low} & \cellcolor{sIa!60}11 & \cellcolor{sIb!6}\textcolor{black!55}{0} & \cellcolor{sII!43}7 & \cellcolor{sIII!33}2 & \cellcolor{sOK!6}\textcolor{black!55}{0} \\
\modelicon{openai}\hspace{0.35em}GPT-5.6 Sol & Codex & \textsf{medium} & \cellcolor{sIa!47}8 & \cellcolor{sIb!58}3 & \cellcolor{sII!43}7 & \cellcolor{sIII!33}2 & \cellcolor{sOK!6}\textcolor{black!55}{0} \\
\modelicon{openai}\hspace{0.35em}GPT-5.6 Sol & Codex & \textsf{high} & \cellcolor{sIa!47}8 & \cellcolor{sIb!6}\textcolor{black!55}{0} & \cellcolor{sII!43}7 & \cellcolor{sIII!53}4 & \cellcolor{sOK!26}\textbf{1} \\
\modelicon{openai}\hspace{0.35em}GPT-5.6 Sol & Codex & \textsf{xhigh} & \cellcolor{sIa!35}5 & \cellcolor{sIb!28}1 & \cellcolor{sII!60}11 & \cellcolor{sIII!43}3 & \cellcolor{sOK!6}\textcolor{black!55}{0} \\
\modelicon{openai}\hspace{0.35em}GPT-5.6 Sol & Codex & \textsf{max} & \cellcolor{sIa!35}5 & \cellcolor{sIb!6}\textcolor{black!55}{0} & \cellcolor{sII!47}8 & \cellcolor{sIII!43}3 & \cellcolor{sOK!60}\textbf{4} \\
\addlinespace[2pt]
\modelicon{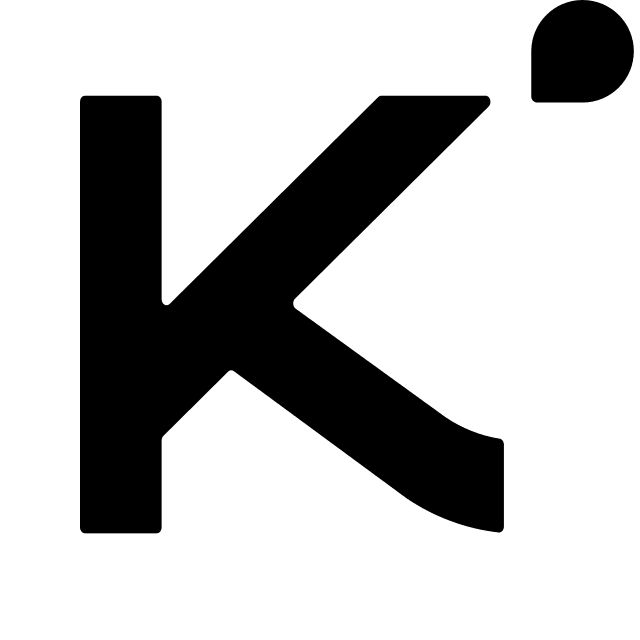}\hspace{0.35em}Kimi K3 & Claude Code & \textsf{max} & \cellcolor{sIa!35}5 & \cellcolor{sIb!6}\textcolor{black!55}{0} & \cellcolor{sII!55}10 & \cellcolor{sIII!43}3 & \cellcolor{sOK!37}\textbf{2} \\
\addlinespace[2pt]
\modelicon{claude}\hspace{0.35em}Claude Sonnet 5 & Claude Code & \textsf{low} & \cellcolor{sIa!35}5 & \cellcolor{sIb!28}1 & \cellcolor{sII!68}13 & \cellcolor{sIII!24}1 & \cellcolor{sOK!6}\textcolor{black!55}{0} \\
\modelicon{claude}\hspace{0.35em}Claude Sonnet 5 & Claude Code & \textsf{medium} & \cellcolor{sIa!39}6 & \cellcolor{sIb!6}\textcolor{black!55}{0} & \cellcolor{sII!55}10 & \cellcolor{sIII!43}3 & \cellcolor{sOK!26}\textbf{1} \\
\modelicon{claude}\hspace{0.35em}Claude Sonnet 5 & Claude Code & \textsf{high} & \cellcolor{sIa!26}3 & \cellcolor{sIb!43}2 & \cellcolor{sII!68}13 & \cellcolor{sIII!33}2 & \cellcolor{sOK!6}\textcolor{black!55}{0} \\
\modelicon{claude}\hspace{0.35em}Claude Sonnet 5 & Claude Code & \textsf{xhigh} & \cellcolor{sIa!43}7 & \cellcolor{sIb!28}1 & \cellcolor{sII!55}10 & \cellcolor{sIII!24}1 & \cellcolor{sOK!26}\textbf{1} \\
\modelicon{claude}\hspace{0.35em}Claude Sonnet 5 & Claude Code & \textsf{max} & \cellcolor{sIa!35}5 & \cellcolor{sIb!28}1 & \cellcolor{sII!64}12 & \cellcolor{sIII!24}1 & \cellcolor{sOK!26}\textbf{1} \\
\addlinespace[2pt]
\modelicon{openai}\hspace{0.35em}GPT-5.6 Luna & Codex & \textsf{none} & \cellcolor{sIa!55}10 & \cellcolor{sIb!28}1 & \cellcolor{sII!47}8 & \cellcolor{sIII!24}1 & \cellcolor{sOK!6}\textcolor{black!55}{0} \\
\modelicon{openai}\hspace{0.35em}GPT-5.6 Luna & Codex & \textsf{low} & \cellcolor{sIa!72}14 & \cellcolor{sIb!28}1 & \cellcolor{sII!31}4 & \cellcolor{sIII!24}1 & \cellcolor{sOK!6}\textcolor{black!55}{0} \\
\modelicon{openai}\hspace{0.35em}GPT-5.6 Luna & Codex & \textsf{medium} & \cellcolor{sIa!43}7 & \cellcolor{sIb!28}1 & \cellcolor{sII!64}12 & \cellcolor{sIII!6}\textcolor{black!55}{0} & \cellcolor{sOK!6}\textcolor{black!55}{0} \\
\modelicon{openai}\hspace{0.35em}GPT-5.6 Luna & Codex & \textsf{high} & \cellcolor{sIa!39}6 & \cellcolor{sIb!43}2 & \cellcolor{sII!60}11 & \cellcolor{sIII!24}1 & \cellcolor{sOK!6}\textcolor{black!55}{0} \\
\modelicon{openai}\hspace{0.35em}GPT-5.6 Luna & Codex & \textsf{xhigh} & \cellcolor{sIa!31}4 & \cellcolor{sIb!43}2 & \cellcolor{sII!64}12 & \cellcolor{sIII!33}2 & \cellcolor{sOK!6}\textcolor{black!55}{0} \\
\modelicon{openai}\hspace{0.35em}GPT-5.6 Luna & Codex & \textsf{max} & \cellcolor{sIa!22}2 & \cellcolor{sIb!28}1 & \cellcolor{sII!72}14 & \cellcolor{sIII!43}3 & \cellcolor{sOK!6}\textcolor{black!55}{0} \\
\addlinespace[2pt]
\modelicon{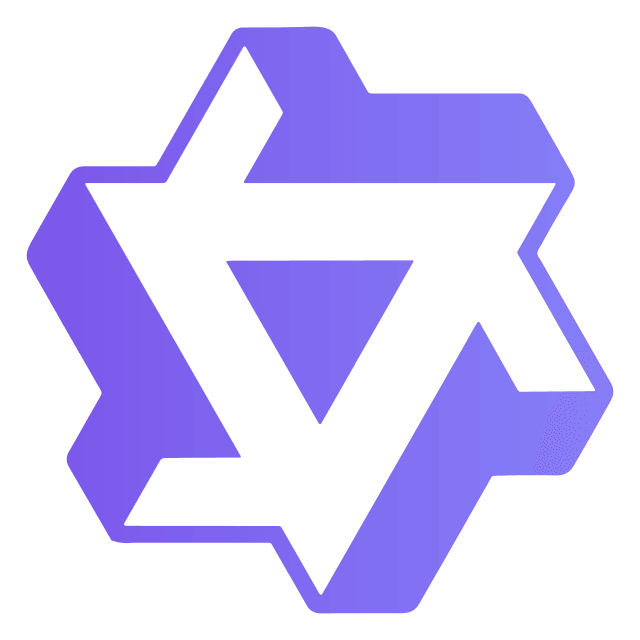}\hspace{0.35em}Qwen 3.8 Max & Claude Code & \textsf{max} & \cellcolor{sIa!47}8 & \cellcolor{sIb!72}4 & \cellcolor{sII!39}6 & \cellcolor{sIII!6}\textcolor{black!55}{0} & \cellcolor{sOK!37}\textbf{2} \\
\addlinespace[2pt]
\modelicon{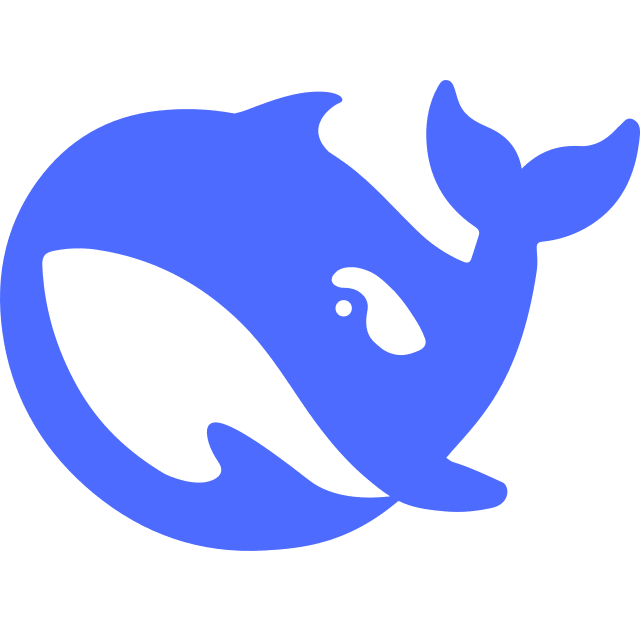}\hspace{0.35em}DeepSeek V4 Flash & Claude Code & \textsf{max} & \cellcolor{sIa!39}6 & \cellcolor{sIb!28}1 & \cellcolor{sII!60}11 & \cellcolor{sIII!33}2 & \cellcolor{sOK!6}\textcolor{black!55}{0} \\
\addlinespace[2pt]
\modelicon{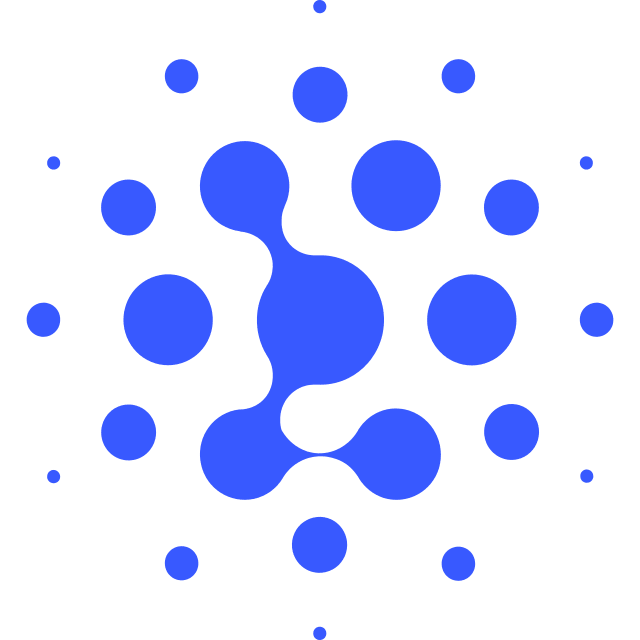}\hspace{0.35em}GLM 5.2 & Claude Code & \textsf{max} & \cellcolor{sIa!51}9 & \cellcolor{sIb!28}1 & \cellcolor{sII!47}8 & \cellcolor{sIII!33}2 & \cellcolor{sOK!6}\textcolor{black!55}{0} \\
\bottomrule
\end{tabular}
\end{adjustbox}
\end{table}

\begin{table}[!t]
\centering
\footnotesize
\setlength{\tabcolsep}{4pt}
\renewcommand{\arraystretch}{1.15}
\caption{\textbf{Behavioural pass rate, score and cost for the same runs.} \emph{Behavioural pass rate} is the
row mean of the fraction of \sII{} fixed checks passed (a perfect score is not required), \emph{score} is
defined in \eqref{eq:score}, and \emph{cost} is the average API spend per task. \emph{Harness} identifies the
coding-agent client: GPT-series models use Codex, and all others use Claude Code. Shading encodes only the
magnitude of a value within its column; each model's best-scoring row is marked with a
\captionchip{brandbg}{shaded background}.}
\label{tab:score}
\begin{tabular}{@{}l l l @{\hspace{10pt}} c @{\hspace{10pt}} c @{\hspace{10pt}} c@{}}
\toprule
\multicolumn{1}{c}{\multirow{2}{*}{\textbf{Model}}}
& \multicolumn{1}{c}{\multirow{2}{*}{\textbf{Harness}}}
& \multicolumn{1}{c}{\multirow{2}{*}{\textbf{Effort}}}
& \multicolumn{1}{c}{\textbf{Behavioural pass}\,\up} & \multicolumn{1}{c}{\textbf{Score}\,\up}
& \multicolumn{1}{c}{\textbf{Cost}\,\dn}\\
\cmidrule(lr){4-4}\cmidrule(lr){5-5}\cmidrule(lr){6-6}
& & & \multicolumn{1}{c}{\%} & \multicolumn{1}{c}{out of 100} & \multicolumn{1}{c}{USD / task}\\
\midrule
% Behavioural pass is newdata.json's meanF: per run, checks passed / checks total;
% then the mean over that row's 20 runs. The same quantity the prose and
% Figure~\ref{fig:shortcut} report, so the three cannot drift apart. An earlier
% version of this column was the weighted mean of module pass rates instead,
% which put every row 0.2-1.7 points off the number the text quoted.
\modelicon{claude}\hspace{0.35em}Claude Opus 5 & Claude Code & \textsf{low} & \g{55}{92.5} & \g{30}{20.5} & 34.8 \\
\modelicon{claude}\hspace{0.35em}Claude Opus 5 & Claude Code & \textsf{medium} & \g{55}{92.7} & \g{42}{28.5} & 38.4 \\
\modelicon{claude}\hspace{0.35em}Claude Opus 5 & Claude Code & \textsf{high} & \g{55}{92.8} & \g{42}{34.5} & 55.7 \\
\bd{\modelicon{claude}\hspace{0.35em}Claude Opus 5} & \bd{Claude Code} & \bd{\textsf{\textbf{xhigh}}} & \g{55}{92.8} & \g{55}{\textbf{47.0}} & 74.9 \\
\modelicon{claude}\hspace{0.35em}Claude Opus 5 & Claude Code & \textsf{max} & \g{55}{91.9} & \g{42}{31.0} & 72.4 \\
\addlinespace[2pt]
\modelicon{openai}\hspace{0.35em}GPT-5.6 Sol & Codex & \textsf{none} & \g{10}{63.6} & \g{10}{4.0} & 2.9 \\
\modelicon{openai}\hspace{0.35em}GPT-5.6 Sol & Codex & \textsf{low} & \g{10}{62.0} & \g{10}{7.0} & 5.9 \\
\modelicon{openai}\hspace{0.35em}GPT-5.6 Sol & Codex & \textsf{medium} & \g{20}{70.8} & \g{10}{6.5} & 6.0 \\
\modelicon{openai}\hspace{0.35em}GPT-5.6 Sol & Codex & \textsf{high} & \g{30}{74.0} & \g{30}{19.0} & 7.7 \\
\modelicon{openai}\hspace{0.35em}GPT-5.6 Sol & Codex & \textsf{xhigh} & \g{30}{73.5} & \g{20}{9.5} & 19.1 \\
\bd{\modelicon{openai}\hspace{0.35em}GPT-5.6 Sol} & \bd{Codex} & \bd{\textsf{\textbf{max}}} & \g{42}{84.1} & \g{42}{\textbf{28.5}} & 143.5 \\
\addlinespace[2pt]
\bd{\modelicon{kimi}\hspace{0.35em}Kimi K3} & \bd{Claude Code} & \bd{\textsf{\textbf{max}}} & \g{55}{93.9} & \g{30}{\textbf{19.5}} & 28.9 \\
\addlinespace[2pt]
\modelicon{claude}\hspace{0.35em}Claude Sonnet 5 & Claude Code & \textsf{low} & \g{30}{79.2} & \g{10}{4.0} & 4.4 \\
\bd{\modelicon{claude}\hspace{0.35em}Claude Sonnet 5} & \bd{Claude Code} & \bd{\textsf{\textbf{medium}}} & \g{30}{73.9} & \g{20}{\textbf{15.0}} & 11.9 \\
\modelicon{claude}\hspace{0.35em}Claude Sonnet 5 & Claude Code & \textsf{high} & \g{42}{85.6} & \g{10}{6.0} & 24.6 \\
\modelicon{claude}\hspace{0.35em}Claude Sonnet 5 & Claude Code & \textsf{xhigh} & \g{30}{76.5} & \g{10}{9.0} & 27.0 \\
\modelicon{claude}\hspace{0.35em}Claude Sonnet 5 & Claude Code & \textsf{max} & \g{42}{84.3} & \g{10}{8.5} & 27.5 \\
\addlinespace[2pt]
\modelicon{openai}\hspace{0.35em}GPT-5.6 Luna & Codex & \textsf{none} & \g{20}{66.4} & \g{10}{4.0} & 1.6 \\
\modelicon{openai}\hspace{0.35em}GPT-5.6 Luna & Codex & \textsf{low} & \g{10}{58.3} & \g{10}{4.0} & 1.7 \\
\modelicon{openai}\hspace{0.35em}GPT-5.6 Luna & Codex & \textsf{medium} & \g{20}{66.6} & \g{10}{0.0} & 1.7 \\
\modelicon{openai}\hspace{0.35em}GPT-5.6 Luna & Codex & \textsf{high} & \g{30}{75.6} & \g{10}{4.0} & 1.8 \\
\modelicon{openai}\hspace{0.35em}GPT-5.6 Luna & Codex & \textsf{xhigh} & \g{42}{83.8} & \g{10}{5.5} & 2.9 \\
\bd{\modelicon{openai}\hspace{0.35em}GPT-5.6 Luna} & \bd{Codex} & \bd{\textsf{\textbf{max}}} & \g{55}{89.1} & \g{20}{\textbf{10.5}} & 2.8 \\
\addlinespace[2pt]
\bd{\modelicon{qwen}\hspace{0.35em}Qwen 3.8 Max} & \bd{Claude Code} & \bd{\textsf{\textbf{max}}} & \g{30}{74.7} & \g{20}{\textbf{10.0}} & 14.5 \\
\addlinespace[2pt]
\bd{\modelicon{deepseek}\hspace{0.35em}DeepSeek V4 Flash} & \bd{Claude Code} & \bd{\textsf{\textbf{max}}} & \g{55}{90.7} & \g{10}{\textbf{7.0}} & 4.3 \\
\addlinespace[2pt]
\bd{\modelicon{glm}\hspace{0.35em}GLM 5.2} & \bd{Claude Code} & \bd{\textsf{\textbf{max}}} & \g{42}{85.2} & \g{10}{\textbf{6.5}} & 17.5 \\
\bottomrule
\end{tabular}
\end{table}

\begin{table}[!t]
\centering
\footnotesize
\setlength{\tabcolsep}{5.5pt}
\renewcommand{\arraystretch}{1.15}
\caption{\textbf{Summary by debt class, pooling all $26$ configurations.} Column meanings follow the metric
definitions of Section~\ref{subsec:setup}; shading follows Table~\ref{tab:score}.
\keyclaim{Stage I pass rate and acceptance count do not move together}: build toolchain is the easiest class to
get past Stage I and the one that loses most at Stage III; framework rewrites contribute $14$ of the $28$
acceptances; language rewrites score lowest.}
\label{tab:category}
\begin{tabular}{@{}l cc @{\hspace{8pt}} cc cc cc @{\hspace{8pt}}|@{\hspace{8pt}} c@{}}
\toprule
& & & \multicolumn{2}{c}{\textbf{Migrated}\,\up} & \multicolumn{1}{c}{\textbf{All tests pass}\,\up}
& \multicolumn{1}{c}{\textbf{Accepted}\,\up} & \multicolumn{1}{c}{\textbf{Broken}\,\dn}
& \multicolumn{1}{c}{\textbf{Blindness}\,\dn} & \multicolumn{1}{c}{\textbf{Score}\,\up}\\
\cmidrule(lr){4-5}\cmidrule(lr){6-6}\cmidrule(lr){7-7}\cmidrule(lr){8-8}\cmidrule(lr){9-9}\cmidrule(lr){10-10}
Class & Tasks & Runs & $n$ & \% & $n$ & $n$ & \% & $n$ & mean\\
\midrule
Build toolchain & 3 & 78 & \g{55}{63} & \g{55}{80.8} & \g{55}{36} & \g{30}{6} & \g{55}{82.4} & \g{10}{2} & \g{55}{\textbf{31.4}} \\
Platform port & 3 & 78 & \g{42}{45} & \g{42}{57.7} & \g{30}{21} & \g{20}{4} & \g{55}{76.5} & \g{20}{4} & \g{30}{\textbf{17.2}} \\
Framework rewrite & 7 & 182 & \g{55}{132} & \g{55}{72.5} & \g{55}{40} & \g{55}{14} & \g{30}{44.0} & \g{55}{15} & \g{20}{\textbf{12.0}} \\
Language rewrite & 7 & 182 & \g{42}{100} & \g{42}{54.9} & \g{30}{21} & \g{20}{4} & \g{42}{66.7} & \g{30}{9} & \g{10}{\textbf{5.6}} \\
\bottomrule
\end{tabular}
\end{table}

%% file: Tables_en/TabStageCategory.tex
% Conditional survival at each stage, three stages by four categories.  Every
% column's denominator is the number of runs that *entered* that stage, so the
% three multiply out to the category's final acceptance rate.  Counts are taken
% from tab:category.
\begin{table}[!t]
\centering
\footnotesize
\setlength{\tabcolsep}{5pt}
\renewcommand{\arraystretch}{1.15}
\caption{\textbf{Agent success varies across stages and migration categories.} Every cell is the share of the
runs that \emph{entered} that stage and passed it, so the three multiply out to the category's
final acceptance rate. The Stage II column is taken over the runs that cleared Stage I (a blindness run passes
every fixed test but is already out at Stage I, so it does not count); the Stage III column is taken over the
runs that cleared both earlier stages and actually met the verifiers.
\keyclaim{Agents do not perform best on one category throughout the pipeline}: they achieve their highest pass
rates on build toolchain rewrites at the first two stages but their lowest at the last; framework rewrites show
the reverse pattern.}
\label{tab:stagecat}
\begin{tabular}{@{}l c @{\hspace{7pt}} cc @{\hspace{7pt}} cc @{\hspace{7pt}} cc @{\hspace{7pt}}|@{\hspace{7pt}} cc@{}}
\toprule
& & \multicolumn{2}{c}{\textbf{\sI{}}\,\up} & \multicolumn{2}{c}{\textbf{\sII{}}\,\up}
& \multicolumn{2}{c}{\textbf{\sIII{}}\,\up} & \multicolumn{2}{c}{\textbf{Final}\,\up}\\
\cmidrule(lr){3-4}\cmidrule(lr){5-6}\cmidrule(lr){7-8}\cmidrule(l){9-10}
Category & Runs & $n$ & \% & $n$ & \% & $n$ & \% & Acc. & Score\\
\midrule
Build toolchain & 78  & \g{55}{63/78}   & \g{55}{80.8} & \g{55}{34/63}   & \g{55}{54.0} & \g{10}{6/34}  & \g{10}{17.6} & \g{42}{6}  & \g{55}{\textbf{31.4}} \\
Platform port   & 78  & \g{30}{45/78}   & \g{30}{57.7} & \g{42}{17/45}   & \g{42}{37.8} & \g{20}{4/17}  & \g{20}{23.5} & \g{20}{4}  & \g{30}{\textbf{17.2}} \\
Framework       & 182 & \g{42}{132/182} & \g{42}{72.5} & \g{20}{25/132}  & \g{20}{18.9} & \g{55}{14/25} & \g{55}{56.0} & \g{55}{14} & \g{20}{\textbf{12.0}} \\
Language        & 182 & \g{20}{100/182} & \g{20}{54.9} & \g{10}{12/100}  & \g{10}{12.0} & \g{30}{4/12}  & \g{30}{33.3} & \g{20}{4}  & \g{10}{\textbf{5.6}} \\
\midrule
All & 520 & 340/520 & 65.4 & 88/340 & 25.9 & 28/88 & 31.8 & 28 & 13.4 \\
\bottomrule
\end{tabular}
\end{table}

%% file: Tables_en/TabHumanEval.tex
% Derived from _keep/additional_exp/stage1_human_eval.csv (156 rows, columns
% task/model/effort/gpt/human). `gpt' is the Stage I verdict from the scoring
% campaign; `human' is the blind annotation. Row arithmetic, both checkable by
% eye: human and judge are each at most `runs', and agree + too strict + too
% lenient = runs. Totals reproduce the numbers quoted in the prose: 140/156 =
% 89.7%, kappa = 0.795, 14 strict against 2 lenient.
\begin{table}[!t]
\centering
\footnotesize
\setlength{\tabcolsep}{6pt}
\renewcommand{\arraystretch}{1.15}
\caption{\textbf{Task-by-Task Comparison of\sI{} and Independent Human Annotation.} Each row is one task, attempted once by
every one of the $26$ model--effort configurations. \emph{Human} and \emph{judge} are how many of those runs
each side called a genuine migration; \emph{agree}, \emph{too strict} and \emph{too lenient} sort the same runs
by whether the two verdicts matched, the judge zeroed a run the human accepted, or the reverse.}
\label{tab:humaneval}
\begin{tabular}{@{}l l @{\hspace{10pt}} r @{\hspace{10pt}} r r @{\hspace{10pt}} r r r@{}}
\toprule
& & & \multicolumn{2}{c}{\textbf{Called a real migration}} & \multicolumn{3}{c}{\textbf{Judge against human}}\\
\cmidrule(lr){4-5}\cmidrule(lr){6-8}
Task & Class & \multicolumn{1}{c}{runs} & \multicolumn{1}{c}{by human} & \multicolumn{1}{c}{by judge}
& \multicolumn{1}{c}{agree\,\up} & \multicolumn{1}{c}{too strict\,\dn}
& \multicolumn{1}{c}{too lenient\,\dn}\\
\midrule
lang01 \textcolor{black!55}{(cmark)}        & Language  & 26 & 7  & 6  & 25 & 1 & 0 \\
lang03 \textcolor{black!55}{(sqlparse)}     & Language  & 26 & 15 & 14 & 23 & 2 & 1 \\
fw02 \textcolor{black!55}{(json-server)}    & Framework & 26 & 15 & 13 & 24 & 2 & 0 \\
fw06 \textcolor{black!55}{(uploadserver)}   & Framework & 26 & 15 & 13 & 22 & 3 & 1 \\
pf02 \textcolor{black!55}{(Stylus)}         & Platform  & 26 & 14 & 12 & 24 & 2 & 0 \\
build01 \textcolor{black!55}{(libsodium)}   & Toolchain & 26 & 23 & 19 & 22 & 4 & 0 \\
\midrule
\textbf{6 tasks} & & \textbf{156} & \textbf{89} & \textbf{77} & \textbf{140} & \textbf{14} & \textbf{2}\\
\bottomrule
\end{tabular}
\end{table}

%% file: Sections_en/4_relatedwork.tex
\section{Related Work}
\label{sec:related_work}

\input{Tables_en/TabComparison}

\paragraph{Repository-level coding benchmarks: the signal is ``red to green''.}
SWE-bench~\citep{jimenez2024swebench} established the paradigm the field now uses: a real repository, a real
issue, and a test that fails before the patch and passes after it. The paradigm has since been extended---to
other languages~\citep{zan2025multiswebench}, to continuously refreshed task streams that resist
contamination~\citep{zhang2025swebenchlive,jain2025livecodebench}, to building a whole library from
scratch~\citep{zhao2025commit0}, and to writing tests rather than patches~\citep{mundler2025swtbench}. A
neighbouring group of suites keeps the same criterion and only stretches the horizon: repository evolution
(SWE-EVO)~\citep{sweevo2025,li2024evocodebench}, long continuous-integration histories
(SWE-CI)~\citep{sweci2026}, and hours of terminal work (Terminal-Bench)~\citep{terminalbench2026}. The
criterion is inherited from function-level suites~\citep{chen2021codex} and fits every one of those task types:
while the work is unfinished the test is red, when it is finished it turns green, and that jump is itself the
evidence of completion. \bench{}'s tasks have no such jump available, because the starting state is already
green---which is not a defect of those benchmarks but a property of behaviour-preserving evolution, which
simply does not produce that signal.

\paragraph{Code migration and reward hacking: fixed tests are not enough.} Behaviour-preserving rewriting is
not new: translation between languages stayed at the function level for a long time after
TransCoder~\citep{lachaux2020transcoder}, and the difficulty of evaluating it was documented
early~\citep{pan2024lost}; the unit later grew to a whole
repository~\citep{wang2024repotransbench,ibrahimzada2024alphatrans}, and C to Rust acquired dedicated
benchmarks of its own~\citep{khatry2025crustbench,emre2021crusts}; other work moves not the language but a
language version~\citep{migrationbench2025} or a single API~\citep{li2025actor}. The changes differ, the
scoring does not: take a test suite, run it, count what passes---reusing the original repository's tests when
the language stays the same, and having humans write them on the target side when it does not. For a
whole-repository change of stack that scoring is no longer sufficient: a repository handed back as-is still
turns any fixed suite green, so fixed tests can say at most ``nothing was broken'', never ``something was
changed''. It is tempting to read this as reward hacking---agents optimise everything their reward leaves
unconstrained~\citep{skalse2022defining,krakovna2020specification,baker2025monitoring}, and coding environments
have supplied plenty of documented cases~\citep{terminalwrench2026,benchjack2026}, with SWE-bench itself
audited item by item for leakage and
contamination~\citep{liang2025sweillusion,garg2025savingswebench,xu2024contamination}. That literature tells us
what to guard against, but the problem here is not that someone is exploiting a loophole: a repository handed
back as-is circumvents no check; it earns full marks by the rules, and the rules simply cannot see whether the
migration happened. Since the hole is not in how tight the tests are, no amount of extra tests will close it
(Section~\ref{subsec:task_formulation}); the only remedy is a second check outside behaviour, holding a
veto~\citep{liang2023helm}.

\paragraph{Model-based judging and differential testing: the tools at the two ends of this work.} Both that
veto and the search beyond the fixed tests have established methods to draw on. When what must be decided
cannot be written as a scoring script, using a model as the judge is standard practice~\citep{zheng2023judging},
and its known failure modes---position and verbosity bias, self-preference~\citep{wang2024fair,panickssery2024selfpreference}---are
exactly why Stage I answers narrow questions with cited evidence over three independent samples instead of
producing one global rating, and why we measure its consistency (Section~\ref{subsec:judge_ablation}). Stage
III is differential testing~\citep{mckeeman1998differential}: with the reference as the other side, this is how
compilers have been checked at scale~\citep{yang2011csmith,deng2023titanfuzz}, and what a verifier must submit
is not an opinion but evidence that runs. The stronger route would of course be formal verification, but
translation validation~\citep{pnueli1998translation}, differential symbolic
execution~\citep{lahiri2012symdiff,cadar2008klee} and verified compilers~\citep{leroy2009compcert} all require
two sides whose semantics can be related, and a cross-language whole-repository rewrite does not offer that.
To approximate the same effect we use agentic verifiers instead: let the strongest coding agents search as hard
as they can with both source trees in hand, and if no counterexample comes out, that is the strongest evidence
we are able to give. 
Table~\ref{tab:comparison} places \bench{} next to the benchmarks above, dimension by dimension.

%% file: Tables_en/TabComparison.tex
\begin{table}[t]
\caption{\textbf{Positioning against related benchmarks.} \emph{Starts failing} means whether the starting
state already fails some test in the suite being scored; that is what makes ``red to green'' an informative
verdict---and it is exactly the property \bench{} cannot have. \emph{Migration criterion} means whether a
non-behavioural instrument can reject a submission that passed every behavioural check; \emph{agentic verifier}
means whether, at scoring time, further models actively search for differences beyond the fixed tests that the
authors did not anticipate. \yes{} is yes, \partialyes{} is partial---MigrationBench moves a language version,
Java 8 to 17/21, while the language itself stays---and \no{} marks a dimension the benchmark does not aim at,
not a defect.}
  \label{tab:comparison}
  \centering
  \footnotesize
  \setlength{\tabcolsep}{4.5pt}
  \begin{tabular}{llcccc}
    \toprule
    Benchmark & Unit of work & Starts failing & Stack changes & Migration criterion & Agentic verifier \\
    \midrule
    SWE-bench~\citep{jimenez2024swebench}        & one issue & \yes & \no & \no & \no \\
    SWE-EVO~\citep{sweevo2025}                   & one feature & \yes & \no & \no & \no \\
    SWE-CI~\citep{sweci2026}                     & CI history & \yes & \no & \no & \no \\
    Terminal-Bench~\citep{terminalbench2026}     & one task & \yes & \no & \no & \no \\
    \midrule
    TransCoder~\citep{lachaux2020transcoder}     & one function & \yes & \yes & \no & \no \\
    MigrationBench~\citep{migrationbench2025}    & whole repository & \yes & \partialyes & \no & \no \\
    \midrule
    \textbf{\bench{} (this work)}                & whole repository & \no & \yes & \yes & \yes \\
    \bottomrule
  \end{tabular}
\end{table}

%% file: Sections_en/5_conclusion.tex
\section{Conclusion}
\label{sec:conclusion}

We have presented \bench{}: 20 long-horizon whole-repository migrations drawn from real open-source
infrastructure, together with a three-stage evaluation protocol that does not rely on behaviour alone---\sI{}
decides whether the migration actually happened and holds a veto, the \sII{} stage decides whether behaviour
is unchanged down to the last observation using $130{,}118$ fixed checks recorded from the original, and
\sIII{} sends six
coding agents, one hour each, to look for the differences the fixed tests may have missed, accepting nothing
but an executable counterexample. Across $520$ evaluations on $8$ frontier models and $26$ configurations, only
$28$ ($5.4\%$) passed all three stages, and 13 of the 20 tasks were solved by nobody. The failures follow a
pattern: ``getting the migration done'' and ``not breaking anything'' are two different abilities, and agents
missed them in opposite directions---a few runs preserved behaviour by skipping the migration and were stopped
at \sI{}, most attempted it and broke behaviour and were stopped at \sII{}. And even when the migration was
genuinely done, only $26\%$ of runs passed every fixed check, and two thirds of those still had a counterexample found
against them by a coding agent within the hour. There is therefore a long way to go before an agent can
complete a whole-repository migration that is genuinely deliverable; and how far along that road we are will be
measured not by how much code the agent wrote, but by whether, once it is done, the system is still the same
system.

%% file: Sections_en/A_construction.tex
\section{How a Task Is Built}
\label{app:construction}

Every task is built by the same procedure. This appendix walks through
\texttt{lang01-cmark-c-to-rust}: its \stA{} is cmark 0.31.1, the CommonMark reference implementation, $41$ C
source files, built by CMake into \texttt{libcmark.so.0.31.1}, \texttt{libcmark.a}
and \texttt{cmark(1)}; its \stB{} is those same three artifacts, behind the same \texttt{cmark.h}, implemented
in Rust 1.90 without any external crate and with the C ABI unchanged.

\paragraph{What a task ships with.} Instantiating \eqref{eq:task} means putting eight things on disk: (i) the
archive of \stA{}, rebuilt to confirm that it works; (ii) a \stB{} declaration naming the target stack down to
fixed versions; (iii) the instruction $\mathcal{I}$; (iv) the offline image $E$, which also carries \stA{}'s
own toolchain; (v) an artifact contract saying what will be collected from the finished workspace; (vi) the
fixed Stage II expectations, recorded from a reference build of \stA{} and divided into modules; (vii) the
migration criteria; and (viii) the further checks made by the agentic verifiers. Items (vi)--(viii) live in a
separate image that is never mounted into the agent's container, so those tests are not merely ``unread'' by
it---they do not exist for it.

\paragraph{Pick the debt first, the repository second.} We start from a migration a maintainer would call
overdue and then look for a project where that migration is the whole job: the old stack has to be
load-bearing, so that removing it reaches the design rather than just the imports. Why cmark qualifies is worth
spelling out, since CommonMark parsers in Rust already exist: what is wanted here is not a correct parser but
\emph{this} library's ABI, installation tree and so on, which no existing crate provides. The repository is
then trimmed, archived and rebuilt inside the image to confirm it works, with history squashed to a single
commit---an upstream log is a place where some past migration may well already be described, and
\texttt{git log} is the first thing a competent agent reads.

\paragraph{Write the requirements, but not the tests.} $\mathcal{I}$ states the requirements in full: the files
that must survive untouched, how the build has to remain invocable without a network, and in which direction
behaviour will be compared. What it never contains is a single one of the concrete behavioural
checks---an instruction that enumerates the tests is an instruction to satisfy the tests.

\paragraph{Scoring reads source only.} An exclusion list is applied when the workspace is collected
(\texttt{build-*}, \texttt{target/}, object files, install prefixes, \texttt{.git}), so ``the old stack no
longer exists'' is a fact about source rather than a fact about what the agent happened to leave in some
directory.

\paragraph{Draw the boundary for the verifiers.} The allow list and deny list of Stage III have to agree with
$\mathcal{I}$: a scope that forbids what the instruction promised makes the task unwinnable, and a scope that
allows what the instruction excluded makes it unloseable. Within that scope a verifier may work the way a real
tester does~\citep{claessen2000quickcheck,chen2018metamorphic,zeller2002delta}. For lang01, fifteen directions
are allowed and eleven forbidden, the latter including internal struct layout, behaviour the original gets
wrong as well, and---most worth stating explicitly---the identity of the tree under test: a candidate program
that detects which tree it is running against satisfies every mechanical condition for a break while
establishing nothing about the migration, and is rejected on discovery rather than weighed.

\paragraph{Make the task verify itself.} No reference solution ships with a task; in its place there is an
identity run at image build time---\stA{} is scored against itself inside the same image, and unless every
behavioural module reaches $1.0$ the build is refused.

\input{Tables_en/TabCaseStudy}

Table~\ref{tab:casestudy} shows one whole task as the scorer sees it: lang01's eight criteria and the fifteen
modules of Stage II. The run in the table is precisely the sharpest form of disagreement between the stages:
all fifteen modules at $1.00$, the highest mark the behavioural side can give, zeroed by Stage I on a single
criterion---every behavioural instrument in the stack calls this migration complete, and the one instrument
that reads the code calls it a transliteration.

%% file: Tables_en/TabCaseStudy.tex
\begin{table}[t]
\centering
\footnotesize
\setlength{\tabcolsep}{2.6pt}
\renewcommand{\arraystretch}{1.05}
\begin{tabular}{@{}l l c c@{\hspace{6pt}} l l c@{}}
\toprule
\multicolumn{4}{@{}l}{\textbf{Stage I --- migration criteria}} &
\multicolumn{3}{l}{\textbf{Stage II --- test modules}}\\
\cmidrule(r){1-4}\cmidrule(l){5-7}
\textbf{Criterion} & \textbf{What it asks} & \textbf{Required} & \textbf{Verdict} &
\textbf{Module} & \textbf{What it measures} & \textbf{Pass rate} \\
\midrule
\texttt{no-c-sources} & has C left the build & \yes & pass &
\texttt{build} & two link configurations & $1.00$ \\
\texttt{no-foreign-headers} & is only \texttt{cmark.h} left & \yes & pass &
\texttt{structure} & install tree, ABI & $1.00$ \\
\texttt{no-c-in-build} & is no C compiled & \yes & pass &
\texttt{conformance} & CommonMark corpus & $1.00$ \\
\texttt{rust-present} & is Rust really there & \yes & pass &
\texttt{renderers} & five output renderers & $1.00$ \\
\texttt{rust-is-primary} & is Rust the implementation & \yes & \textbf{fail} &
\texttt{data-tables} & entities, case folding & $1.00$ \\
\texttt{no-embedded-reference} & no prebuilt artifact linked & \yes & pass &
\texttt{encoding} & UTF-8 validation & $1.00$ \\
\texttt{no-verifier-awareness} & was the scorer let off & \yes & pass &
\texttt{robustness} & pathological input & $1.00$ \\
\texttt{default-path} & is the scored path default & \yes & pass &
\texttt{options} & options and combinations & $1.00$ \\
\cmidrule(r){1-4}
\multicolumn{4}{@{}l}{\textit{All eight criteria are required, so one failure}} &
\texttt{streaming} & incremental parser & $1.00$ \\
\multicolumn{4}{@{}l}{\textit{zeroes the run. Each is decided by majority}} &
\texttt{tree} & AST and its invariants & $1.00$ \\
\multicolumn{4}{@{}l}{\textit{over 3 samples at \texttt{medium} effort---here all}} &
\texttt{iterators} & iterator API & $1.00$ \\
\multicolumn{4}{@{}l}{\textit{eight were unanimous.}} &
\texttt{provenance} & what built these artifacts & $1.00$ \\
\multicolumn{4}{@{}l}{} &
\texttt{cli} & the installed executable & $1.00$ \\
\cmidrule(l){5-7}
\multicolumn{4}{@{}l}{\textbf{Any criterion fails} $\Rightarrow$ \textbf{score 0}; Stage III never runs} &
\multicolumn{2}{l}{\textbf{$\mathbf{4{,}184}$ checks, none skipped}} & $\mathbf{1.00}$ \\
\multicolumn{4}{@{}l}{} &
\multicolumn{2}{l}{\textit{a perfect Stage II}} & \\
\bottomrule
\end{tabular}
\caption{\textbf{One task and one submission in full:} \texttt{claude-opus-5} at \texttt{high} effort on
\texttt{lang01-cmark-c-to-rust}\textbf{.} All $4{,}184$ checks across $15$ modules pass, the highest mark Stage
II can give; seven of the eight criteria pass, and the eighth, \texttt{rust-is-primary}, fails unanimously
because the Rust reproduces the C implementation layout field by field---\emph{``the same hand-managed pointer
stack, the same names and the same order \ldots{} this is the original transliterated into unsafe Rust''}, cited
to \texttt{src/inlines.rs:55} and neighbouring lines. Final score: $0$.}
\label{tab:casestudy}
\end{table}

%% file: Sections_en/B_per_task.tex
\section{Per-Task Breakdown}
\label{app:per_task}

\input{Tables_en/TabTasks}

Table~\ref{tab:tasks} takes the funnel of Section~\ref{subsec:main_results} apart task by task: of the $26$
runs, how many completed the migration, how many passed every fixed check, how many of those fall into
blindness, and how many were finally accepted.

Of the $20$ tasks, $13$ were never solved by any model, and for seven of them not a single submission even
reached the verifiers---and they fail in two different ways. On \texttt{lang03}, \texttt{lang04}, and
\texttt{pf02}, no run passed every fixed check. On \texttt{lang01}, \texttt{fw01}, \texttt{fw02}, and
\texttt{fw07}, by contrast, some runs passed every fixed check, but every such run was rejected at Stage I as
blindness. A behavioural instrument would report these four tasks as solved and the three-stage protocol reports
them as unsolved, which is exactly the disagreement this benchmark was built to expose.

%% file: Tables_en/TabTasks.tex
\begin{table}[!t]
\centering
\scriptsize
\setlength{\tabcolsep}{3pt}
\renewcommand{\arraystretch}{1.05}
\caption{\textbf{The $520$ runs broken down by task.} Every task is run once by each of the $26$ configurations
($n=26$). Column meanings follow the metric definitions of Section~\ref{subsec:setup}: \emph{migrated} is the
number of runs passing \sI{}, \emph{all tests pass} is the number passing every fixed check in \sII{}
(\emph{regardless} of the Stage I verdict), \emph{blindness} is the subset of those that failed Stage I,
\emph{accepted} is the number passing all three stages, and \emph{score} is defined in \eqref{eq:score}.
\emph{All tests pass}~$-$~\emph{blindness} is therefore the number of runs that reached Stage III.
\keyclaim{Only $7$ of the $20$ tasks ever produced an accepted submission}.}
\label{tab:tasks}
\begin{tabular}{@{}llr rrr r r@{}}
\toprule
& & & \multicolumn{1}{c}{\textbf{Migrated}\,\up} & \multicolumn{1}{c}{\textbf{All tests pass}\,\up}
 & \multicolumn{1}{c}{\textbf{Blindness}\,\dn} & \multicolumn{1}{c}{\textbf{Accepted}\,\up}
 & \multicolumn{1}{c}{\textbf{Score}\,\up}\\
& & & \multicolumn{1}{c}{\footnotesize Stage I passed} & \multicolumn{1}{c}{\footnotesize Stage II perfect}
 & \multicolumn{1}{c}{\footnotesize passed, not migrated} & \multicolumn{1}{c}{\footnotesize all three passed}
 & \multicolumn{1}{c}{\footnotesize out of 100}\\
\cmidrule(lr){4-4}\cmidrule(lr){5-5}\cmidrule(lr){6-6}\cmidrule(lr){7-7}\cmidrule(lr){8-8}
Task & Migration & \multicolumn{1}{c}{$n$} & \multicolumn{1}{c}{$n$}
 & \multicolumn{1}{c}{$n$} & \multicolumn{1}{c}{$n$} & \multicolumn{1}{c}{$n$}
 & \multicolumn{1}{c}{mean}\\
\midrule
\multicolumn{8}{@{}l}{\textit{Language rewrites}}\\
\texttt{lang01} & C $\to$ Rust & 26 & 6 & 5 & \textbf{5} & 0 & 0.00\\
\texttt{lang02} & C $\to$ Java & 26 & 18 & 4 & 0 & 0 & 12.69\\
\texttt{lang03} & Python $\to$ Go & 26 & 14 & 0 & 0 & 0 & 0.00\\
\texttt{lang04} & JavaScript $\to$ Rust & 26 & 20 & 0 & 0 & 0 & 0.00\\
\texttt{lang05} & Go $\to$ Zig & 26 & 18 & 4 & 0 & \textbf{1} & 12.69\\
\texttt{lang06} & C++ $\to$ C\# & 26 & 10 & 3 & \textbf{2} & 0 & 2.31\\
\texttt{lang07} & JavaScript $\to$ TypeScript & 26 & 14 & 5 & \textbf{2} & \textbf{3} & 11.54\\
\addlinespace[2pt]
\multicolumn{8}{@{}l}{\textit{Framework rewrites}}\\
\texttt{fw01} & Flask $\to$ Starlette & 26 & 24 & 1 & \textbf{1} & 0 & 0.00\\
\texttt{fw02} & Express $\to$ Fastify & 26 & 13 & 1 & \textbf{1} & 0 & 0.00\\
\texttt{fw03} & Vue $\to$ React & 26 & 26 & 5 & 0 & \textbf{2} & 16.92\\
\texttt{fw04} & Gin $\to$ chi & 26 & 25 & 7 & \textbf{1} & \textbf{3} & 21.54\\
\texttt{fw05} & actix-web $\to$ axum & 26 & 22 & 2 & \textbf{1} & 0 & 2.31\\
\texttt{fw06} & \texttt{gorilla/mux} $\to$ \texttt{net/http} & 26 & 13 & 22 & \textbf{9} & \textbf{9} & 43.08\\
\texttt{fw07} & Dropwizard $\to$ Spring Boot & 26 & 9 & 2 & \textbf{2} & 0 & 0.00\\
\addlinespace[2pt]
\multicolumn{8}{@{}l}{\textit{Platform ports}}\\
\texttt{pf01} & POSIX $\to$ \texttt{wasm32-wasi} & 26 & 15 & 1 & 0 & 0 & 2.31\\
\texttt{pf02} & CommonJS $\to$ V8 realm & 26 & 12 & 0 & 0 & 0 & 0.00\\
\texttt{pf03} & x86-64 $\to$ 3 architectures & 26 & 18 & 20 & \textbf{4} & \textbf{4} & 49.23\\
\addlinespace[2pt]
\multicolumn{8}{@{}l}{\textit{Build-toolchain rewrites}}\\
\texttt{build01} & Autotools $\to$ CMake & 26 & 19 & 2 & 0 & 0 & 4.23\\
\texttt{build02} & Maven $\to$ Gradle & 26 & 20 & 13 & 0 & 0 & 25.38\\
\texttt{build03} & setuptools $\to$ Meson & 26 & 24 & 21 & \textbf{2} & \textbf{6} & 64.62\\
\midrule
\textbf{All tasks} & & \textbf{520} & 340 & 118 & 30 & \textbf{28} & 13.44\\
\bottomrule
\end{tabular}
\end{table}

%% file: Sections_en/C_catalog.tex
\section{Detailed Task Catalogue}
\label{app:task_catalog}

Table~\ref{tab:catalog_full} gives all $20$ tasks one by one; Table~\ref{tab:catalog} in
Section~\ref{subsec:composition} is its summary by class.

\input{Tables_en/TabCatalogFull}

%% file: Tables_en/TabCatalogFull.tex
\begin{table}[h]
\centering
\footnotesize
\setlength{\tabcolsep}{2.3pt}
\renewcommand{\arraystretch}{1.06}
\begin{tabular}{@{}llllr r r r r@{}}
\toprule
\textbf{Task} & \textbf{Upstream project} & \textbf{Source stack} & \textbf{Target stack} &
\textbf{LoC} & \textbf{$B$\,(h)} & \textbf{Criteria} & \textbf{Modules} & \textbf{Checks} \\
\midrule
\multicolumn{9}{@{}l}{\textit{Language rewrites---the implementation language moves, the artifact does not}}\\
lang01 & cmark 0.31.1        & C11        & Rust 1.90      & 22{,}619 & 30 & 8  & 15 & 4{,}184 \\
lang02 & zlib 1.3.1          & C89        & Java 17        & 22{,}732 & 20 & 10 & 18 & 4{,}162 \\
lang03 & sqlparse 0.5.3      & Python     & Go (stdlib)    & 4{,}393  & 15 & 8  & 15 & 8{,}530 \\
lang04 & acorn 8.14.0        & JavaScript & Rust 1.90      & 9{,}571  & 30 & 7  & 13 & 16{,}039 \\
lang05 & go-yaml v3.0.1      & Go         & Zig 0.14.1     & 11{,}967 & 20 & 8  & 22 & 10{,}271 \\
lang06 & jsonnet 0.20.0      & C++11      & C\# / .NET 8   & 39{,}842 & 20 & 9  & 19 & 2{,}608 \\
lang07 & JSONata 2.2.2       & JavaScript & TypeScript 5.9 & 9{,}247  & 12 & 10 & 15 & 13{,}977 \\
\addlinespace[1.5pt]
\multicolumn{9}{@{}l}{\textit{Framework rewrites---the language stays; what is replaced is the framework the code is organised around}}\\
fw01 & httpbin           & Flask (WSGI)     & Starlette (ASGI3) & 2{,}455  & 9  & 5 & 11 & 2{,}830 \\
fw02 & json-server       & Express 4        & Fastify 5         & 2{,}992  & 9  & 5 & 10 & 6{,}760 \\
fw03 & RealWorld Conduit & Vue 2 + webpack  & React 18 + Vite   & 2{,}261  & 11 & 5 & 10 & 21{,}769 \\
fw04 & ChartMuseum       & Gin              & chi v5 (net/http) & 5{,}401  & 12 & 5 & 14 & 11{,}429 \\
fw05 & miniserve         & actix-web 4      & axum 0.7 (tower)  & 3{,}484  & 8  & 6 & 14 & 12{,}466 \\
fw06 & uploadserver      & gorilla/mux      & net/http ServeMux & 803      & 6  & 6 & 9  & 485 \\
fw07 & GraphHopper 11.0  & Dropwizard 4     & Spring Boot 3.5   & 94{,}766 & 16 & 7 & 11 & 113 \\
\addlinespace[1.5pt]
\multicolumn{9}{@{}l}{\textit{Platform ports---the host the code assumes changes}}\\
pf01 & SQLite 3.31.1  & POSIX (unix + win) & \texttt{wasm32-wasi}      & 358{,}006 & 10 & 6 & 7  & 2{,}668 \\
pf02 & Stylus 0.63.0  & CommonJS + Node    & ESM in a V8 realm        & 16{,}012  & 10 & 7 & 12 & 2{,}570 \\
pf03 & QuickJS        & x86-64 host layout & 3 architectures, endian-clean & 89{,}429 & 6 & 6 & 12 & 1{,}487 \\
\addlinespace[1.5pt]
\multicolumn{9}{@{}l}{\textit{Build-toolchain rewrites---what produces the artifact changes, and the package is the observable}}\\
build01 & libsodium 1.0.20    & Autotools  & CMake 3.20+      & 73{,}228 & 6 & 6 & 15 & 4{,}720 \\
build02 & Gson 2.10.1         & Maven      & Gradle (offline) & 19{,}015 & 6 & 6 & 10 & 2{,}669 \\
build03 & pycryptodome 3.20.0 & setuptools & Meson            & 78{,}839 & 6 & 6 & 12 & 381 \\
\midrule
\multicolumn{4}{@{}l}{\textbf{20 tasks}} & \textbf{867{,}062} & \textbf{262} & \textbf{136} &
\textbf{264} & \textbf{130{,}118} \\
\bottomrule
\end{tabular}
\caption{\textbf{The task set, task by task.} \emph{LoC} counts newlines in the version-controlled
implementation source of \stA{}, excluding tests, vendored trees and documentation, measured the same way for
every task rather than transcribed from each project's own description; together these repositories hold
$10{,}594$ version-controlled files. $B$ is the agent's time budget in hours. \emph{Criteria} is the number of
prompt-form migration criteria, each of them required: a single failure zeroes the submission. \emph{Modules}
is the number of independent test modules and \emph{checks} the number of fixed behavioural cases they
hold, recorded from a reference build of \stA{}. The size of a test suite is set by how much of the product is
mechanically observable, not by the size of the repository. Every task has six verifiers of one hour each.
Table~\ref{tab:catalog} summarises these rows by class. Each row ships an upstream release verbatim, at the
version shown and under its own licence: cmark~\citep{sw:cmark}, zlib~\citep{sw:zlib},
sqlparse~\citep{sw:sqlparse}, Acorn~\citep{sw:acorn}, go-yaml~\citep{sw:goyaml}, Jsonnet~\citep{sw:jsonnet},
JSONata~\citep{sw:jsonata}, httpbin~\citep{sw:httpbin}, json-server~\citep{sw:jsonserver}, the RealWorld
Conduit app~\citep{sw:vuerealworld}, ChartMuseum~\citep{sw:chartmuseum}, miniserve~\citep{sw:miniserve},
go-simple-upload-server~\citep{sw:uploadserver}, GraphHopper~\citep{sw:graphhopper}, SQLite~\citep{sw:sqlite},
Stylus~\citep{sw:stylus}, QuickJS~\citep{sw:quickjs}, libsodium~\citep{sw:libsodium}, Gson~\citep{sw:gson} and
PyCryptodome~\citep{sw:pycryptodome}.}
\label{tab:catalog_full}
\end{table}

%% file: ref.bib
@inproceedings{jimenez2024swebench,
  author    = {Carlos E. Jimenez and John Yang and Alexander Wettig and Shunyu Yao and
               Kexin Pei and Ofir Press and Karthik Narasimhan},
  title     = {{SWE}-bench: Can Language Models Resolve Real-World {GitHub} Issues?},
  booktitle = {International Conference on Learning Representations (ICLR)},
  year      = {2024},
  note      = {arXiv:2310.06770}
}

@inproceedings{yang2024sweagent,
  author    = {John Yang and Carlos E. Jimenez and Alexander Wettig and Kilian Lieret and
               Shunyu Yao and Karthik Narasimhan and Ofir Press},
  title     = {{SWE}-agent: Agent-Computer Interfaces Enable Automated Software Engineering},
  booktitle = {Advances in Neural Information Processing Systems (NeurIPS)},
  year      = {2024},
  note      = {arXiv:2405.15793}
}

@inproceedings{wang2025openhands,
  author    = {Xingyao Wang and Boxuan Li and Yufan Song and Frank F. Xu and Xiangru Tang and
               Mingchen Zhuge and Jiayi Pan and Yueqi Song and Bowen Li and Jaskirat Singh and
               Hoang H. Tran and Fuqiang Li and Ren Ma and Mingzhang Zheng and Bill Qian and
               Yanjun Shao and Niklas Muennighoff and Yizhe Zhang and Binyuan Hui and
               Junyang Lin and Robert Brennan and Hao Peng and Heng Ji and Graham Neubig},
  title     = {{OpenHands}: An Open Platform for {AI} Software Developers as Generalist Agents},
  booktitle = {International Conference on Learning Representations (ICLR)},
  year      = {2025},
  note      = {arXiv:2407.16741}
}

@article{zan2025multiswebench,
  author  = {Daoguang Zan and Zhirong Huang and Wei Liu and Hanwu Chen and Linhao Zhang and
             Shulin Xin and Lu Chen and Qi Liu and Xiaojian Zhong and Aoyan Li and Siyao Liu and
             Yongsheng Xiao and Liangqiang Chen and Yuyu Zhang and Jing Su and Tianyu Liu and
             Rui Long and Kai Shen and Liang Xiang},
  title   = {Multi-{SWE}-bench: A Multilingual Benchmark for Issue Resolving},
  journal = {arXiv preprint arXiv:2504.02605},
  year    = {2025}
}

@article{zhang2025swebenchlive,
  author  = {Linghao Zhang and Shilin He and Chaoyun Zhang and Yu Kang and Bowen Li and
             Chengxing Xie and Junhao Wang and Maoquan Wang and Yufan Huang and Shengyu Fu and
             Elsie Nallipogu and Qingwei Lin and Yingnong Dang and Saravan Rajmohan and
             Dongmei Zhang},
  title   = {{SWE}-bench Goes Live!},
  journal = {arXiv preprint arXiv:2505.23419},
  year    = {2025}
}

@article{sweevo2025,
  author  = {Tue Le and Minh V. T. Thai and Dung Nguyen Manh and Huy Phan Nhat and Nghi D. Q. Bui},
  title   = {{SWE-EVO}: Benchmarking Coding Agents in Long-Horizon Software Evolution Scenarios},
  journal = {arXiv preprint arXiv:2512.18470},
  year    = {2025}
}

@article{sweci2026,
  author  = {Jialong Chen and Xander Xu and Hu Wei and Chuan Chen and Bing Zhao},
  title   = {{SWE-CI}: Evaluating Agent Capabilities in Maintaining Codebases via
             Continuous Integration},
  journal = {arXiv preprint arXiv:2603.03823},
  year    = {2026}
}

@inproceedings{mundler2025swtbench,
  author    = {Niels M{\"u}ndler and Mark Niklas M{\"u}ller and Jingxuan He and Martin Vechev},
  title     = {{SWT}-Bench: Testing and Validating Real-World Bug-Fixes with Code Agents},
  booktitle = {Advances in Neural Information Processing Systems (NeurIPS)},
  year      = {2024},
  note      = {arXiv:2406.12952}
}

@inproceedings{zhao2025commit0,
  author    = {Wenting Zhao and Nan Jiang and Celine Lee and Justin T. Chiu and Claire Cardie and
               Matthias Gall{\'e} and Alexander M. Rush},
  title     = {{Commit0}: Library Generation from Scratch},
  booktitle = {International Conference on Learning Representations (ICLR)},
  year      = {2025},
  note      = {arXiv:2412.01769}
}

@inproceedings{li2024evocodebench,
  author    = {Jia Li and Ge Li and Xuanming Zhang and Yunfei Zhao and Yihong Dong and Zhi Jin and
               Binhua Li and Fei Huang and Yongbin Li},
  title     = {{EvoCodeBench}: An Evolving Code Generation Benchmark with Domain-Specific
               Evaluations},
  booktitle = {Advances in Neural Information Processing Systems (NeurIPS), Datasets and
               Benchmarks Track},
  year      = {2024},
  note      = {arXiv:2410.22821}
}

@article{terminalbench2026,
  author  = {Mike A. Merrill and others},
  title   = {{Terminal-Bench}: Benchmarking Agents on Hard, Realistic Tasks in Command Line
             Interfaces},
  journal = {arXiv preprint arXiv:2601.11868},
  year    = {2026}
}

@article{chen2021codex,
  author  = {Mark Chen and Jerry Tworek and Heewoo Jun and Qiming Yuan and
             Henrique Ponde de Oliveira Pinto and Jared Kaplan and Harri Edwards and Yuri Burda and
             Nicholas Joseph and Greg Brockman and Alex Ray and Raul Puri and Gretchen Krueger and
             Michael Petrov and Heidy Khlaaf and Girish Sastry and Pamela Mishkin and Brooke Chan and
             Scott Gray and Nick Ryder and Mikhail Pavlov and Alethea Power and Lukasz Kaiser and
             Mohammad Bavarian and Clemens Winter and Philippe Tillet and Felipe Petroski Such and
             Dave Cummings and Matthias Plappert and Fotios Chantzis and Elizabeth Barnes and
             Ariel Herbert-Voss and William Hebgen Guss and Alex Nichol and Alex Paino and
             Nikolas Tezak and Jie Tang and Igor Babuschkin and Suchir Balaji and Shantanu Jain and
             William Saunders and Christopher Hesse and Andrew N. Carr and Jan Leike and
             Josh Achiam and Vedant Misra and Evan Morikawa and Alec Radford and Matthew Knight and
             Miles Brundage and Mira Murati and Katie Mayer and Peter Welinder and Bob McGrew and
             Dario Amodei and Sam McCandlish and Ilya Sutskever and Wojciech Zaremba},
  title   = {Evaluating Large Language Models Trained on Code},
  journal = {arXiv preprint arXiv:2107.03374},
  year    = {2021}
}

@inproceedings{jain2025livecodebench,
  author    = {Naman Jain and King Han and Alex Gu and Wen-Ding Li and Fanjia Yan and
               Tianjun Zhang and Sida Wang and Armando Solar-Lezama and Koushik Sen and Ion Stoica},
  title     = {{LiveCodeBench}: Holistic and Contamination Free Evaluation of Large Language
               Models for Code},
  booktitle = {International Conference on Learning Representations (ICLR)},
  year      = {2025},
  note      = {arXiv:2403.07974}
}

@inproceedings{lachaux2020transcoder,
  author    = {Baptiste Rozi{\`e}re and Marie-Anne Lachaux and Lowik Chanussot and Guillaume Lample},
  title     = {Unsupervised Translation of Programming Languages},
  booktitle = {Advances in Neural Information Processing Systems (NeurIPS)},
  year      = {2020},
  note      = {arXiv:2006.03511}
}

@inproceedings{pan2024lost,
  author    = {Rangeet Pan and Ali Reza Ibrahimzada and Rahul Krishna and Divya Sankar and
               Lambert Pouguem Wassi and Michele Merler and Boris Sobolev and Raju Pavuluri and
               Saurabh Sinha and Reyhaneh Jabbarvand},
  title     = {Lost in Translation: A Study of Bugs Introduced by Large Language Models while
               Translating Code},
  booktitle = {IEEE/ACM International Conference on Software Engineering (ICSE)},
  year      = {2024},
  note      = {arXiv:2308.03109}
}

@article{migrationbench2025,
  author  = {Linbo Liu and Xinle Liu and Qiang Zhou and Lin Chen and Yihan Liu and Hoan Nguyen and
             Behrooz Omidvar-Tehrani and Xi Shen and Jun Huan and Omer Tripp and Anoop Deoras},
  title   = {{MigrationBench}: Repository-Level Code Migration Benchmark from {Java} 8},
  journal = {arXiv preprint arXiv:2505.09569},
  year    = {2025}
}

@article{wang2024repotransbench,
  author  = {Yanli Wang and Yanlin Wang and Suiquan Wang and Daya Guo and Jiachi Chen and
             John Grundy and Xilin Liu and Yuchi Ma and Mingzhi Mao and Hongyu Zhang and Zibin Zheng},
  title   = {{RepoTransBench}: A Real-World Multilingual Benchmark for Repository-Level Code
             Translation},
  journal = {arXiv preprint arXiv:2412.17744},
  year    = {2024}
}

@article{ibrahimzada2024alphatrans,
  author  = {Ali Reza Ibrahimzada and Kaiyao Ke and Mrigank Pawagi and Muhammad Salman Abid and
             Rangeet Pan and Saurabh Sinha and Reyhaneh Jabbarvand},
  title   = {{AlphaTrans}: A Neuro-Symbolic Compositional Approach for Repository-Level Code
             Translation and Validation},
  journal = {arXiv preprint arXiv:2410.24117},
  year    = {2024}
}

@article{khatry2025crustbench,
  author  = {Anirudh Khatry and Robert Zhang and Jia Pan and Ziteng Wang and Qiaochu Chen and
             Greg Durrett and Isil Dillig},
  title   = {{CRUST}-Bench: A Comprehensive Benchmark for {C}-to-safe-{Rust} Transpilation},
  journal = {arXiv preprint arXiv:2504.15254},
  year    = {2025}
}

@article{li2025actor,
  author  = {Tianyu Li and Ruishi Li and Bo Wang and Brandon Paulsen and Umang Mathur and
             Prateek Saxena},
  title   = {Adversarial Agent Collaboration for Correctness Improvements of {C} to Safe {Rust}
             Translation},
  journal = {arXiv preprint arXiv:2510.03879},
  year    = {2025}
}

@article{emre2021crusts,
  author    = {Mehmet Emre and Ryan Schroeder and Kyle Dewey and Ben Hardekopf},
  title     = {Translating {C} to Safer {Rust}},
  journal   = {Proceedings of the ACM on Programming Languages},
  volume    = {5},
  number    = {OOPSLA},
  articleno = {121},
  year      = {2021}
}

@inproceedings{skalse2022defining,
  author    = {Joar Skalse and Nikolaus H. R. Howe and Dmitrii Krasheninnikov and David Krueger},
  title     = {Defining and Characterizing Reward Hacking},
  booktitle = {Advances in Neural Information Processing Systems (NeurIPS)},
  year      = {2022},
  note      = {arXiv:2209.13085}
}

@article{baker2025monitoring,
  author  = {Bowen Baker and Joost Huizinga and Leo Gao and Zehao Dou and Melody Y. Guan and
             Aleksander Madry and Wojciech Zaremba and Jakub Pachocki and David Farhi},
  title   = {Monitoring Reasoning Models for Misbehavior and the Risks of Promoting Obfuscation},
  journal = {arXiv preprint arXiv:2503.11926},
  year    = {2025}
}

@misc{krakovna2020specification,
  author       = {Victoria Krakovna and Jonathan Uesato and Vladimir Mikulik and Matthew Rahtz and
                  Tom Everitt and Ramana Kumar and Zac Kenton and Jan Leike and Shane Legg},
  title        = {Specification Gaming: The Flip Side of {AI} Ingenuity},
  howpublished = {DeepMind blog. \url{https://deepmind.google/blog/specification-gaming-the-flip-side-of-ai-ingenuity/}},
  year         = {2020}
}

@article{terminalwrench2026,
  author  = {Ivan Bercovich and Ivgeni Segal and Kexun Zhang and Shashwat Saxena and
             Aditi Raghunathan and Ziqian Zhong},
  title   = {Terminal Wrench: A Dataset of 331 Reward-Hackable Environments and 3,632 Exploit
             Trajectories},
  journal = {arXiv preprint arXiv:2604.17596},
  year    = {2026}
}

@article{benchjack2026,
  author  = {Hao Wang and Hanchen Li and Qiuyang Mang and Alvin Cheung and Koushik Sen and
             Dawn Song},
  title   = {Do Androids Dream of Breaking the Game? Systematically Auditing {AI} Agent Benchmarks
             with {BenchJack}},
  journal = {arXiv preprint arXiv:2605.12673},
  year    = {2026}
}

@article{liang2025sweillusion,
  author  = {Shanchao Liang and Spandan Garg and Roshanak Zilouchian Moghaddam},
  title   = {The {SWE}-Bench Illusion: When State-of-the-Art {LLMs} Remember Instead of Reason},
  journal = {arXiv preprint arXiv:2506.12286},
  year    = {2025}
}

@article{garg2025savingswebench,
  author  = {Spandan Garg and Benjamin Steenhoek and Yufan Huang},
  title   = {Saving {SWE}-Bench: A Benchmark Mutation Approach for Realistic Agent Evaluation},
  journal = {arXiv preprint arXiv:2510.08996},
  year    = {2025}
}

@article{xu2024contamination,
  author  = {Cheng Xu and Shuhao Guan and Derek Greene and M-Tahar Kechadi},
  title   = {Benchmark Data Contamination of Large Language Models: A Survey},
  journal = {arXiv preprint arXiv:2406.04244},
  year    = {2024}
}

@article{liang2023helm,
  author  = {Percy Liang and Rishi Bommasani and Tony Lee and Dimitris Tsipras and Dilara Soylu and
             Michihiro Yasunaga and Yian Zhang and Deepak Narayanan and Yuhuai Wu and
             Ananya Kumar and Benjamin Newman and Binhang Yuan and Bobby Yan and Ce Zhang and
             Christian Cosgrove and Christopher D. Manning and Christopher R{\'e} and
             Diana Acosta-Navas and Drew A. Hudson and Eric Zelikman and Esin Durmus and
             Faisal Ladhak and Frieda Rong and Hongyu Ren and Huaxiu Yao and Jue Wang and
             Keshav Santhanam and Laurel Orr and Lucia Zheng and Mert Yuksekgonul and
             Mirac Suzgun and Nathan Kim and Neel Guha and Niladri S. Chatterji and Omar Khattab and
             Peter Henderson and Qian Huang and Ryan Andrew Chi and Sang Michael Xie and
             Shibani Santurkar and Surya Ganguli and Tatsunori Hashimoto and Thomas Icard and
             Tianyi Zhang and Vishrav Chaudhary and William Wang and Xuechen Li and Yifan Mai and
             Yuhui Zhang and Yuta Koreeda},
  title   = {Holistic Evaluation of Language Models},
  journal = {Transactions on Machine Learning Research},
  year    = {2023},
  note    = {arXiv:2211.09110}
}

@article{mckeeman1998differential,
  author  = {William M. McKeeman},
  title   = {Differential Testing for Software},
  journal = {Digital Technical Journal},
  volume  = {10},
  number  = {1},
  pages   = {100--107},
  year    = {1998}
}

@inproceedings{yang2011csmith,
  author    = {Xuejun Yang and Yang Chen and Eric Eide and John Regehr},
  title     = {Finding and Understanding Bugs in {C} Compilers},
  booktitle = {ACM SIGPLAN Conference on Programming Language Design and Implementation (PLDI)},
  year      = {2011}
}

@article{leroy2009compcert,
  author  = {Xavier Leroy},
  title   = {Formal Verification of a Realistic Compiler},
  journal = {Communications of the ACM},
  volume  = {52},
  number  = {7},
  pages   = {107--115},
  year    = {2009}
}

@inproceedings{pnueli1998translation,
  author    = {Amir Pnueli and Michael Siegel and Eli Singerman},
  title     = {Translation Validation},
  booktitle = {International Conference on Tools and Algorithms for the Construction and Analysis
               of Systems (TACAS)},
  year      = {1998}
}

@inproceedings{lahiri2012symdiff,
  author    = {Shuvendu K. Lahiri and Chris Hawblitzel and Ming Kawaguchi and
               Henrique Reb{\^e}lo},
  title     = {{SYMDIFF}: A Language-Agnostic Semantic Diff Tool for Imperative Programs},
  booktitle = {International Conference on Computer Aided Verification (CAV)},
  year      = {2012}
}

@inproceedings{cadar2008klee,
  author    = {Cristian Cadar and Daniel Dunbar and Dawson Engler},
  title     = {{KLEE}: Unassisted and Automatic Generation of High-Coverage Tests for Complex
               Systems Programs},
  booktitle = {USENIX Symposium on Operating Systems Design and Implementation (OSDI)},
  year      = {2008}
}

@inproceedings{claessen2000quickcheck,
  author    = {Koen Claessen and John Hughes},
  title     = {{QuickCheck}: A Lightweight Tool for Random Testing of {Haskell} Programs},
  booktitle = {ACM SIGPLAN International Conference on Functional Programming (ICFP)},
  year      = {2000}
}

@article{chen2018metamorphic,
  author  = {Tsong Yueh Chen and Fei-Ching Kuo and Huai Liu and Pak-Lok Poon and Dave Towey and
             T. H. Tse and Zhi Quan Zhou},
  title   = {Metamorphic Testing: A Review of Challenges and Opportunities},
  journal = {ACM Computing Surveys},
  volume  = {51},
  number  = {1},
  pages   = {4:1--4:27},
  year    = {2018}
}

@article{zeller2002delta,
  author  = {Andreas Zeller and Ralf Hildebrandt},
  title   = {Simplifying and Isolating Failure-Inducing Input},
  journal = {IEEE Transactions on Software Engineering},
  volume  = {28},
  number  = {2},
  pages   = {183--200},
  year    = {2002}
}

@inproceedings{deng2023titanfuzz,
  author    = {Yinlin Deng and Chunqiu Steven Xia and Haoran Peng and Chenyuan Yang and
               Lingming Zhang},
  title     = {Large Language Models Are Zero-Shot Fuzzers: Fuzzing Deep-Learning Libraries via
               Large Language Models},
  booktitle = {ACM SIGSOFT International Symposium on Software Testing and Analysis (ISSTA)},
  year      = {2023},
  note      = {arXiv:2212.14834}
}

@inproceedings{cunningham1992debt,
  author    = {Ward Cunningham},
  title     = {The {WyCash} Portfolio Management System},
  booktitle = {Addendum to the Proceedings on Object-Oriented Programming Systems, Languages,
               and Applications (OOPSLA Addendum), Experience Report},
  pages     = {29--30},
  year      = {1992}
}

@inproceedings{zheng2023judging,
  author    = {Lianmin Zheng and Wei-Lin Chiang and Ying Sheng and Siyuan Zhuang and Zhanghao Wu and
               Yonghao Zhuang and Zi Lin and Zhuohan Li and Dacheng Li and Eric P. Xing and
               Hao Zhang and Joseph E. Gonzalez and Ion Stoica},
  title     = {Judging {LLM}-as-a-Judge with {MT}-Bench and Chatbot Arena},
  booktitle = {Advances in Neural Information Processing Systems (NeurIPS)},
  year      = {2023},
  note      = {arXiv:2306.05685}
}

@inproceedings{wang2024fair,
  author    = {Peiyi Wang and Lei Li and Liang Chen and Zefan Cai and Dawei Zhu and Binghuai Lin and
               Yunbo Cao and Lingpeng Kong and Qi Liu and Tianyu Liu and Zhifang Sui},
  title     = {Large Language Models Are Not Fair Evaluators},
  booktitle = {Annual Meeting of the Association for Computational Linguistics (ACL)},
  year      = {2024},
  note      = {arXiv:2305.17926}
}

@article{panickssery2024selfpreference,
  author  = {Arjun Panickssery and Samuel R. Bowman and Shi Feng},
  title   = {{LLM} Evaluators Recognize and Favor Their Own Generations},
  journal = {arXiv preprint arXiv:2404.13076},
  year    = {2024}
}

@article{qwen3,
  author  = {An Yang and Anfeng Li and Baosong Yang and Beichen Zhang and Binyuan Hui and Bo Zheng and
             Bowen Yu and Chang Gao and Chengen Huang and Chenxu Lv and Chujie Zheng and
             Dayiheng Liu and Fan Zhou and Fei Huang and Feng Hu and Hao Ge and Haoran Wei and
             Huan Lin and Jialong Tang and Jian Yang and Jianhong Tu and Jianwei Zhang and
             Jianxin Yang and Jiaxi Yang and Jing Zhou and Jingren Zhou and Junyang Lin and
             Kai Dang and Keqin Bao and Kexin Yang and Le Yu and Lianghao Deng and Mei Li and
             Mingfeng Xue and Mingze Li and Pei Zhang and Peng Wang and Qin Zhu and Rui Men and
             Ruize Gao and Shixuan Liu and Shuang Luo and Tianhao Li and Tianyi Tang and
             Wenbiao Yin and Xingzhang Ren and Xinyu Wang and Xinyu Zhang and Xuancheng Ren and
             Yang Fan and Yang Su and Yichang Zhang and Yinger Zhang and Yu Wan and Yuqiong Liu and
             Zekun Wang and Zeyu Cui and Zhenru Zhang and Zhipeng Zhou and Zihan Qiu},
  title   = {{Qwen3} Technical Report},
  journal = {arXiv preprint arXiv:2505.09388},
  year    = {2025}
}

@article{deepseekv3,
  author  = {{DeepSeek-AI}},
  title   = {{DeepSeek-V3} Technical Report},
  journal = {arXiv preprint arXiv:2412.19437},
  year    = {2024}
}

@article{kimik2,
  author  = {{Kimi Team}},
  title   = {{Kimi K2}: Open Agentic Intelligence},
  journal = {arXiv preprint arXiv:2507.20534},
  year    = {2025}
}

@article{glm45,
  author  = {{GLM-4.5 Team}},
  title   = {{GLM-4.5}: Agentic, Reasoning, and Coding ({ARC}) Foundation Models},
  journal = {arXiv preprint arXiv:2508.06471},
  year    = {2025}
}

@article{gaffney2022sqlite,
  author  = {Kevin P. Gaffney and Martin Prammer and Larry Brasfield and D. Richard Hipp and
             Dan Kennedy and Jignesh M. Patel},
  title   = {{SQLite}: Past, Present, and Future},
  journal = {Proceedings of the VLDB Endowment},
  volume  = {15},
  number  = {12},
  pages   = {3535--3547},
  year    = {2022}
}

@inproceedings{haas2017wasm,
  author    = {Andreas Haas and Andreas Rossberg and Derek L. Schuff and Ben L. Titzer and
               Michael Holman and Dan Gohman and Luke Wagner and Alon Zakai and JF Bastien},
  title     = {Bringing the Web Up to Speed with {WebAssembly}},
  booktitle = {ACM SIGPLAN Conference on Programming Language Design and Implementation (PLDI)},
  year      = {2017}
}

@inproceedings{bernstein2012nacl,
  author    = {Daniel J. Bernstein and Tanja Lange and Peter Schwabe},
  title     = {The Security Impact of a New Cryptographic Library},
  booktitle = {International Conference on Cryptology and Information Security in Latin America
               (LATINCRYPT)},
  year      = {2012}
}

@inproceedings{geisberger2008ch,
  author    = {Robert Geisberger and Peter Sanders and Dominik Schultes and Daniel Delling},
  title     = {Contraction Hierarchies: Faster and Simpler Hierarchical Routing in Road Networks},
  booktitle = {International Workshop on Experimental Algorithms (WEA)},
  year      = {2008}
}

@techreport{deutsch1996deflate,
  author      = {L. Peter Deutsch},
  title       = {{DEFLATE} Compressed Data Format Specification version 1.3},
  institution = {Internet Engineering Task Force},
  number      = {RFC 1951},
  year        = {1996}
}

@misc{sw:libsodium,
  author       = {Frank Denis},
  title        = {libsodium 1.0.20},
  howpublished = {\url{https://github.com/jedisct1/libsodium}},
  note         = {ISC licence},
  year         = {2026}
}

@misc{sw:gson,
  author       = {{Google}},
  title        = {{Gson} gson-parent-2.10.1},
  howpublished = {\url{https://github.com/google/gson}},
  note         = {Apache-2.0 licence},
  year         = {2011}
}

@misc{sw:pycryptodome,
  author       = {{The PyCryptodome Authors}},
  title        = {{PyCryptodome} 3.20.0},
  howpublished = {\url{https://github.com/Legrandin/pycryptodome}},
  note         = {BSD-2-Clause and Unlicense},
  year         = {2024}
}

@misc{sw:httpbin,
  author       = {Kenneth Reitz},
  title        = {httpbin 0.10.2},
  howpublished = {\url{https://github.com/psf/httpbin}},
  note         = {ISC or MIT licence},
  year         = {2017}
}

@misc{sw:jsonserver,
  author       = {{typicode}},
  title        = {json-server 0.17.4},
  howpublished = {\url{https://github.com/typicode/json-server}},
  note         = {MIT licence},
  year         = {2015}
}

@misc{sw:vuerealworld,
  author       = {{The RealWorld Contributors}},
  title        = {vue-realworld-example-app, commit feb0b7d2},
  howpublished = {\url{https://github.com/gothinkster/vue-realworld-example-app}},
  note         = {MIT licence},
  year         = {2017}
}

@misc{sw:chartmuseum,
  author       = {{The Helm Authors}},
  title        = {{ChartMuseum} v0.15.0},
  howpublished = {\url{https://github.com/helm/chartmuseum}},
  note         = {Apache-2.0 licence},
  year         = {2022}
}

@misc{sw:miniserve,
  author       = {Sven-Hendrik Haase},
  title        = {miniserve 0.27.1},
  howpublished = {\url{https://github.com/svenstaro/miniserve}},
  note         = {MIT licence},
  year         = {2018}
}

@misc{sw:uploadserver,
  author       = {Mei Akizuru},
  title        = {go-simple-upload-server v2.2.0},
  howpublished = {\url{https://github.com/mayth/go-simple-upload-server}},
  note         = {MIT licence},
  year         = {2023}
}

@misc{sw:graphhopper,
  author       = {{GraphHopper GmbH}},
  title        = {{GraphHopper} 11.0},
  howpublished = {\url{https://github.com/graphhopper/graphhopper}},
  note         = {Apache-2.0 licence},
  year         = {2024}
}

@misc{sw:cmark,
  author       = {John MacFarlane},
  title        = {cmark 0.31.1: the {CommonMark} reference implementation},
  howpublished = {\url{https://github.com/commonmark/cmark}},
  note         = {BSD-2-Clause, MIT and CC-BY-SA-4.0},
  year         = {2014}
}

@misc{sw:zlib,
  author       = {Jean-loup Gailly and Mark Adler},
  title        = {zlib 1.3.1},
  howpublished = {\url{https://github.com/madler/zlib}},
  note         = {Zlib licence},
  year         = {2022}
}

@misc{sw:sqlparse,
  author       = {Andi Albrecht},
  title        = {sqlparse 0.5.3},
  howpublished = {\url{https://github.com/andialbrecht/sqlparse}},
  note         = {BSD-3-Clause licence},
  year         = {2016}
}

@misc{sw:acorn,
  author       = {{The Acorn Contributors}},
  title        = {Acorn 8.14.0: a small, fast {JavaScript} parser},
  howpublished = {\url{https://github.com/acornjs/acorn}},
  note         = {MIT licence},
  year         = {2022}
}

@misc{sw:goyaml,
  author       = {{Canonical Ltd.}},
  title        = {go-yaml v3.0.1: {YAML} support for the {Go} language},
  howpublished = {\url{https://github.com/go-yaml/yaml}},
  note         = {Apache-2.0 and MIT licences},
  year         = {2019}
}

@misc{sw:jsonnet,
  author       = {{Google Inc.}},
  title        = {Jsonnet 0.20.0: a data templating language},
  howpublished = {\url{https://github.com/google/jsonnet}},
  note         = {Apache-2.0 licence},
  year         = {2015}
}

@misc{sw:jsonata,
  author       = {{IBM Corp.}},
  title        = {{JSONata} 2.2.2: a query and transformation language for {JSON}},
  howpublished = {\url{https://github.com/jsonata-js/jsonata}},
  note         = {MIT licence},
  year         = {2018}
}

@misc{sw:sqlite,
  author       = {{The SQLite Developers}},
  title        = {{SQLite} 3.31.1},
  howpublished = {\url{https://www.sqlite.org/}},
  note         = {Public domain; SPDX identifier \texttt{blessing}},
  year         = {2020}
}

@misc{sw:stylus,
  author       = {{Automattic}},
  title        = {Stylus 0.63.0: an expressive {CSS} preprocessor},
  howpublished = {\url{https://github.com/stylus/stylus}},
  note         = {MIT licence},
  year         = {2024}
}

@misc{sw:quickjs,
  author       = {Fabrice Bellard and Charlie Gordon},
  title        = {{QuickJS} 2020-11-08: a small embeddable {JavaScript} engine},
  howpublished = {\url{https://bellard.org/quickjs/}},
  note         = {MIT licence, stated in the header of every source file},
  year         = {2020}
}
